\documentclass[twoside]{article}

\usepackage[accepted]{aistats2024}
\usepackage{booktabs}       % professional-quality tables
\usepackage{amsfonts}       % blackboard math symbols
\usepackage{url}
\usepackage{nicefrac}       % compact symbols for 1/2, etc.
\usepackage{microtype}      % microtypography
\usepackage[dvipsnames]{xcolor}        % colors
\usepackage{graphicx}
\usepackage{graphics}
\usepackage{amsmath}
\usepackage{multirow}
\usepackage{amsthm}
\usepackage{bbm}

\usepackage{soul}
\usepackage{subfigure}
\usepackage{subcaption}
\usepackage{mathabx}
\usepackage[round]{natbib}
\usepackage{comment}

\newcommand{\re}[1]{{\textcolor{black}
{#1}}}
\newcommand{\zs}[1]{{\textcolor{black}
{#1}}}
\newcommand{\oldrevise}[1]{{\textcolor{black}
{#1}}}
\begin{document}



\twocolumn[

\aistatstitle{Non-Neighbors Also Matter to Kriging: A New Contrastive-Prototypical Learning}

\aistatsauthor{Zhishuai Li\textsuperscript{1}, Yunhao Nie\textsuperscript{1,2}, Ziyue Li\textsuperscript{3}, Lei Bai\textsuperscript{4} , Yisheng Lv\textsuperscript{2}, Rui Zhao\textsuperscript{1} }
\aistatsaddress{ \textsuperscript{1}SenseTime Research,  \textsuperscript{2}Chinese Academy of Sciences, \textsuperscript{3}University of Cologne, \textsuperscript{4}Shanghai AI Laboratory}
 ]

\begin{abstract}
Kriging aims to estimate the attributes of unseen geo-locations from observations in the spatial vicinity or physical connections. %, which helps mitigate skewed monitoring caused by under-deployed sensors. 
Existing works assume that neighbors' information offers the basis for estimating the unobserved target while ignoring non-neighbors. 
However, neighbors could also be \re{quite different or even misleading}, and the non-neighbors could still offer constructive information. To this end, we propose “\textbf{C}ontrastive-\textbf{P}rototypical” self-supervised learning for \textbf{K}riging (KCP): %to refine valuable information from neighbors and recycle the one from non-neighbors. %Rather than directly predicting the attributes of unobserved nodes, 
% As a pre-trained paradigm, we conduct the Kriging task from a new perspective of representation: we aim to first learn robust and general representations and then recover attributes from representations. %since they are usually more robust and general than the raw input.
(1) The neighboring contrastive module coarsely pushes neighbors together and non-neighbors apart. %learns the representations by narrowing the representation distance between the target and its neighbors while pushing away the non-neighbors. 
(2) In parallel, the prototypical module identifies similar representations via exchanged prediction, such that it refines the misleading neighbors and recycles the useful non-neighbors from the neighboring contrast component. As a result, \textit{not all} the neighbors and \textit{some} of the non-neighbors will be used to infer the target.
(3) To learn general and robust representations, we design an adaptive augmentation module that encourages data diversity. \re{Theoretical bound is derived for the proposed augmentation.} %incorporates data-driven attribute augmentation and centrality-based topology augmentation over the spatiotemporal Kriging graph data. %This module , facilitating more accurate attribute estimation. 
%Then, in collaboration with the graph neural network backbone, %we design a neighboring contrastive module that coarsely learns the representations by narrowing the representation distance between the target and its neighbors while pushing away the non-neighbors. 
%Further, a prototypical module is introduced to identify similar representations via exchanged prediction regardless of the spatial adjacency, thus refining the misleading neighbors and recycling the useful non-neighbors from the neighboring contrast component. 
Extensive experiments on real-world datasets demonstrate the superior performance of KCP compared to its peers with \textbf{6\%} improvements and exceptional transferability and robustness. The code is available at \url{https://github.com/bonaldli/KCP}.
%The code shall be released upon acceptance.
\end{abstract}

\section{Introduction}
% where to make a stand: 
% - Kriging (primary)
% - spatiotemporal SSL (secondary, general, and task-agnostic)

% introduction story
% - motivation of Kriging: sensor coverage rate
% - current methods and their limitations (assumption: using neighbours)
\oldrevise{Spatially-distributed sensors are commonly deployed to perceive the environment, such as temperature-humidity sensors in weather stations in meteorology \citep{gad2017performance}, landside tilt sensors in geology \citep{li2021multi}, water-level sensors in hydrogeology \citep{tonkin2002kriging}, geomagnetic sensors \citep{kwong2009arterial} or cameras \citep{lin2021vehicle,han2022dr} in transportation, and so on.}
% Those ubiquitous sensing technologies have profound impacts on the sustainable development of society. 
\begin{figure}[t]
    \centering
    \includegraphics[width=1\columnwidth]{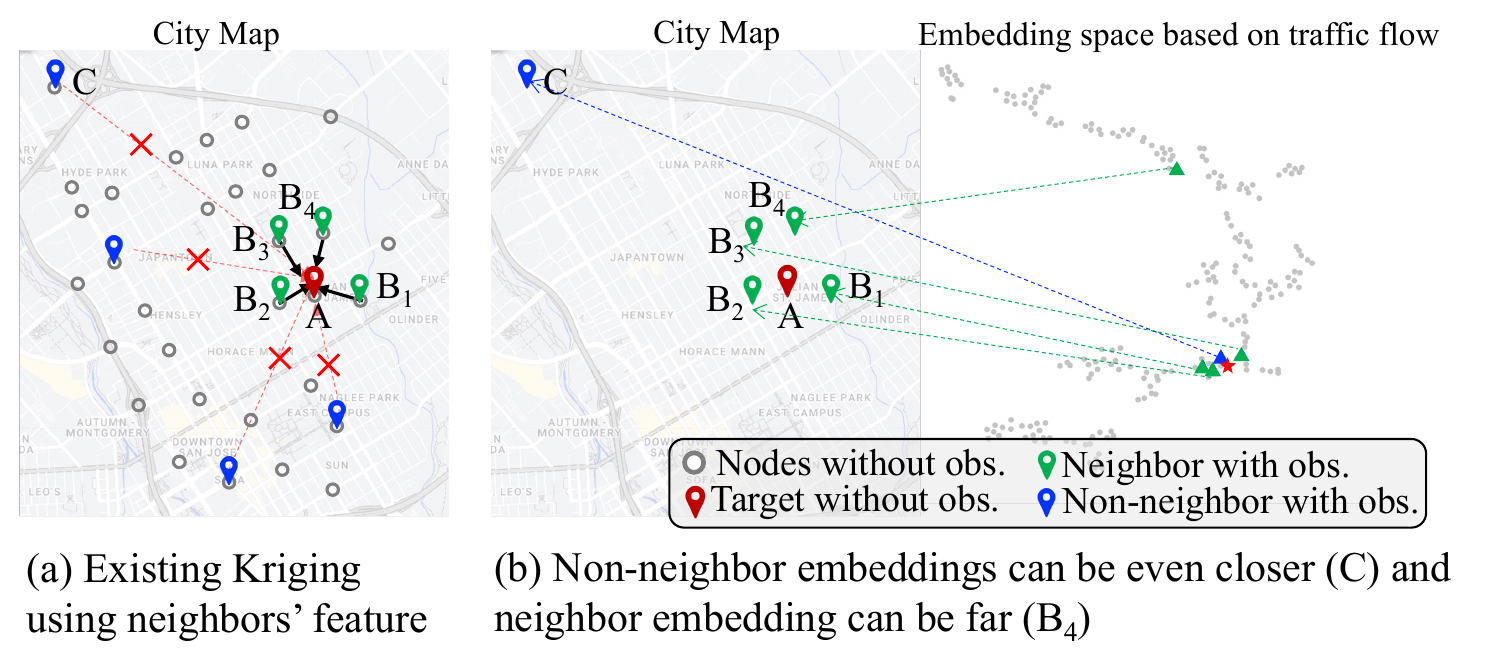}
    \caption{When looking through the node representations in the embedding space, spatial neighbors do not always appear nearby and non-neighbors may also be close to the target. %\oldrevise{add Fig. 1c: targets and neighbors attributes}
    }
    \label{fig:embedding}
        \vspace{-15pt}
\end{figure}

However, it is prohibitively costly to deploy sensors with high coverage rates, resulting in under-sampling and skewed monitoring. \oldrevise{For example, in the traffic domain, some cities only have less than 20\% intersections installed with the sensors to collect the traffic flow. To leverage data-driven applications, the data collected from spatially sporadic sensors must be expanded into spatially fine-grained data.} 
Tailored to such a spatial super-resolution task, \textit{Kriging}, also known as \textit{spatial interpolation}, %aims at 
infers the attributes of targets at unsampled locations from observations in the spatial vicinity or physical connections. \oldrevise{Shown in Fig. \ref{fig:embedding}(a), Kriging exploits a few locations' observed data (blue/green pins) to infer the rest that is not even equipped with sensors (grey circles).}  

\oldrevise{\textit{\textbf{``Common Practice'' of Kriging: Neighbors only.}} Existing Kriging models \citep{bostan2017basic, appleby2020kriging, wu2021inductive, wu2021spatial, lei2022bayesian} are built upon the assumption} that the information of neighbors offers the basis to estimate the attribute values of the target. As shown in Fig. \ref{fig:embedding}(a), closer neighbor nodes (e.g., green nodes $B_1$ to $B_4$) have larger impacts on the interpolation weight of the target node (e.g., the red node $A$), but the non-neighbors (e.g., the blue nodes) should be ignored, \re{called  ``law of neighborhood''.} \oldrevise{To achieve so, statistical Kriging introduces graph Laplacian penalty \citep{rao2015collaborative}, and deep Kriging based on Graph Convolution Networks (GCN) \citep{wu2021inductive, appleby2020kriging} also relies on adjacency matrix, pushing neighbors' embeddings closer.} %%%%%%%%%%\oldrevise{However, if the work uses a general correlation matrix \citep{lei2022bayesian} as a penalty or GAT, then do they also manage to capture non-neighbor information? how to justify our difference compared with multi-view GCN etc that also consider non-physical neighbors}
%\textcolor{red}{$\blacktriangleup$}

\oldrevise{\textbf{\textit{A New Perspective: Solely relying on the neighborhood is not always error-proof. Representation speaks louder than neighbors.}} Here we take a closer look at Fig. \ref{fig:embedding}(b):} we conduct graph encoding via GCN \citep{kipf2016semi} based on node attribute - traffic flow, and visualize the embeddings via t-SNE \citep{van2008visualizing}, with the target A as red \textcolor{red}{$\star$}, its neighbors $B_1$ to $B_4$ as green \textcolor{Green}{$\triangle$}, and its non-neighbor, e.g., $C$, as blue \textcolor{blue}{$\triangle$}. 
The insights are two-fold: (1) the red \textcolor{red}{$\star$} is close to the most of green \textcolor{Green}{$\triangle$}: it aligns with the general assumption - neighbors have similar representations in embedding space. (2) Some non-neighbors' embeddings can be even closer despite not being physical neighbors, e.g., the blue \textcolor{blue}{$\triangle$} is even closer; \oldrevise{some neighbor, i.e., $B_4$ yet has quite a different embedding.} %(3) in the experiment, we also discovered that neighbors could also offer nonconstructive or misleading information. 
Inspired by the context, we rethink Kriging from a novel perspective: \oldrevise{\textbf{to interpolate by using the embeddings rather than the raw input}. Specifically, we learn the target embedding first and then recover the values of unknown targets via downstream modules. %It is worth mentioning that, outside the scope of Kriging, in general spatiotemporal analysis, there are methods that explicitly construct other semantic graphs to utilize non-neighbors' information. Still, this explicit design requires domain knowledge, and these methods do not have the \textit{inductive} ability to learn the nodes that do not appear during the training. Details will be discussed in Section 2.
} % and then recovering the values of unknown nodes by the downstream module, since the embedding can preserve receptive properties across the graph and considerably overcome the effects of scattered noise.

\oldrevise{To conclude, we will formulate the Kriging task in a pre-training and finetuning paradigm. We try to solve the following three research questions (RQ):}
\\{\textbf{RQ1:}}\label{RQ1} How to first conform to the \textit{common practice of Kriging}, \zs{that is, estimating the target node with its neighbors and improving the aggregation effecicy}?
\\{\textbf{RQ2:}}\label{RQ2} On the top of \textbf{RQ1}, how to further utilize the non-neighbors' useful information and discard the neighbors' nonconstructive information and maintain the balance of the two (some may use more no-neighbors and others may use more neighbors)?
\\{\textbf{RQ3:}}\label{RQ3} For representation learning, how to learn a more general embedding that respects the spatiotemporal nature of the data, reflects the Kriging task, and is also potentially robust enough to noise?

To answer all the questions, we propose a \textbf{K}riging \textbf{C}ontrastive-\textbf{P}rototypical Learning (KCP). \oldrevise{Since selecting which nodes to infer the target is based on representation, high-quality embeddings are now the first key, while the second key is how to select the rational nodes. Self-supervised learning (SSL) has proved its superiority in learning general embedding, so this work will be also the first Kriging solution based on contrastive learning}. 
Three critical modules are designed, shown in Fig. \ref{fig:model}: The contrastive module respects the ``common practice'' by attracting the neighbors' embeddings together and repelling non-neighbors' embeddings away. \zs{It can elaborate the embedding of the target node by its neighbors, which answers \textbf{RQ1}; the prototypical module instead learns to discard the neighbors' nonconstructive embeddings (\textit{refine}) and keep the non-neighbors' useful embeddings (\textit{recycle}) via exchanged prediction, which responses \textbf{RQ2}. 
% The combination of contrastive and prototypical learning streak a delicate balance between the old norm and the new perspective, answering \textbf{RQ1} and \textbf{RQ2} respectively. 
To get a robust representation, the third component, a spatiotemporal adaptive augmentation module, is proposed, which augments the input graph from (spatial) topology and (temporal) feature views, in an adaptive and probabilistic manner to diversify the augmented data, answering the \textbf{RQ3}.}

The main contributions are summarized as follows:

 (1) To the best of our knowledge, we are the first to break the common practice of Kriging, i.e., neighbors only, and propose a novel self-supervised learning framework to better aggregate information from \textit{not all} neighbors and \textit{some} non-neighbors for Kriging.
 
 (2)  We let the two SSL modules, i.e., neighboring contrast and prototypical head, collaborate %for effective supervision 
    \oldrevise{to refine and recycle constructive information for target nodes}. \oldrevise{To facilitate more robust representation learning, we also propose an adaptive augmentation module to generate diverse and Kriging-related data for the SSL modules.} %All of them can take full use of neighboring and no-neighboring information.
  
    (3) We conduct extensive experiments on three real-world datasets to evaluate the superiority of the proposed KCP under various settings, which achieves $ 3\% \sim 6\%$ improvements over its peers and demonstrates the best transferability and robustness.
\vspace{-4pt}
\section{Related work}
% \oldrevise{better to categorize them as: transductive v.s. inductive; statistical methods v.s. deep methods}
\vspace{-4pt}
%\subsection{Kriging methods}
% The fundamental intuition behind Kriging is to model the spatial correlation across observed points to estimate the attributes at unobserved locations \citep{krige1951statistical,goovaerts1998ordinary}. 
%The taxonomy of 
Kriging methods can be categorized as \textit{transductive} and \textit{inductive}. The details can be referred to in Supplementary Materials 1.% (1) \oldrevise{Transductive models require all the nodes to be present during training. %and it cannot learn the representation for the unseen nodes in a natural way: %the graph in testing not to exceed the scope of the observed graph in training. 
\vspace{-4pt}
\subsection{Inductive Kriging}
\vspace{-4pt}
The message-passing mechanism in graph neural networks (GNNs), %\oldrevise{not all GNNs are inductive?}, 
such as GraphSAGE \citep{hamilton2017inductive}, makes them well-suited for inductive Kriging, in that they can effectively aggregate the helpful information to evaluate unknown data points.
By predefining and investigating the $K$ nearest neighbors around the unobserved node, KCN \citep{appleby2020kriging} estimates the targets by averaging the neighbors' labels with learnable weights. 
\zs{Furthermore, with randomly selected $K$ neighbors, PE-GNN \citep{klemmer2023positional} plugs a general and highly modular positional encoding component to learn the context-aware embedding for geographic coordinates.
By the modified Moran's I auxiliary task, the module can be well-trained in parallel with the main task,  incorporating spatial context and correlation explicitly.
Similarly, \cite{egressy2022graph} finds that positional node embeddings derived from the coordinate position under the stress function can be very effective for graph-based applications, and can learn to generalize with limited training data. These provide novel insights into graph-based Kriging.}
IGNNK \citep{wu2021inductive} generates random subgraphs and reconstructs the signals on them (especially on unobserved nodes) by learning the spatial correlations.
\zs{But it may lack the capture of node-level temporal dependencies.} Aiming at better information aggregation from neighbors, SATCN \citep{wu2021spatial} designs several node message-passing modules for graph learning and mines temporal correlations through a temporal convolutional network. \zs{INCREASE \citep{zheng2023increase} explicitly defines geographic and semantic neighbors for the target node, thus providing additionally referred candidates. But it may also introduce more false positive neighbors as confused terms. Overall, the key insight of GNNs for Kriging is how to improve the information aggregation efficiency, \oldrevise{especifically spatiotemporal correlations from either neighbors or no-neighbors}, which relates to the model's robustness and inductive ability.}
% \vspace{-4pt}
% \subsection{Non-Neighbors}\vspace{-4pt} %It is worth mentioning that, 
% In general spatiotemporal analysis, there are methods that explicitly construct other semantic graphs to utilize non-neighbors' information. Still, this explicit design requires domain knowledge, and these methods are not \textit{inductive}. % to learn the unseen nodes that do not appear during the training. 
% Details are in Suppl. 1.2.
%\textbf{Methods that Consider Non-neighbors.} Outside the scope of Kriging, there are methods that also consider the nodes that are not physical neighbors. They mainly introduce more semantic graphs such as Point of Interest (POI) \citep{li2020tensor}, transition \citep{mao2022jointly}, and connectivity \citep{geng2019spatiotemporal}, besides the topological graph. These semantic graphs are either incorporated as additional graph Laplacian penalties in statistical models such as matrix factorization \citep{yang2021real} and tensor decomposition \citep{li2020tensor}, or constructed as multi-view graphs in GCN-based models \citep{geng2019spatiotemporal}. However, these explicitly defined graphs require domain knowledge, which might not be wholesome to explain why two nodes are similar. We prefer to select similar or dissimilar nodes in a learning base. Moreover, these methods are transductive.
\zs{\subsection{Graph structure learning on road network}
Pioneering work, such as STGCN \citep{yu2018spatio} and DCRNN \citep{li2018diffusion}, emphasizes the inherent topology of the road network, such as a binary adjacency graph, or utilizes predefined graphs based on specific metrics like Euclidean distance to indicate the graph structure. Tailored to the traffic data input, GWNet \citep{wu2019graph} proposes the learnable embedding metric for pairwise node distance construction, which automatically constructs adaptive graphs for road networks. According to the node embeddings, AGCRN \citep{bai2020adaptive} introduces node-specific convolution filters to infer the inter-dependencies among different traffic series automatically.
Rather than the Euclidean distance metric in road works, MFFB \citep{li2020multi} proposes evaluating the node distance by Spearman similarity with trainable bias, thus building a dynamic adaptive graph.
MTGODE \citep{jin2022multivariate} abstracts multivariate time series into dynamic graphs with time-evolving node features and unknown graph structures. Based on the formulation, it designs and solves a neural ODE to complement missing graph topologies and unify both spatial and temporal message passing. These studies enhance the feasibility of building adaptive and general graphs but cannot be directly applied to the Kriging task, since we only recognize the basic attributes of unseen nodes and lack any historical observations, while historical data is indispensable in adaptive graph structure learning.}
\vspace{-4pt}
\section{Methodology}
\vspace{-4pt}
% In this section, we will provide the problem formulation and then elaborate on the proposed KCP. 
\begin{figure*}[t]
    \centering
    \includegraphics[width=0.99\textwidth]{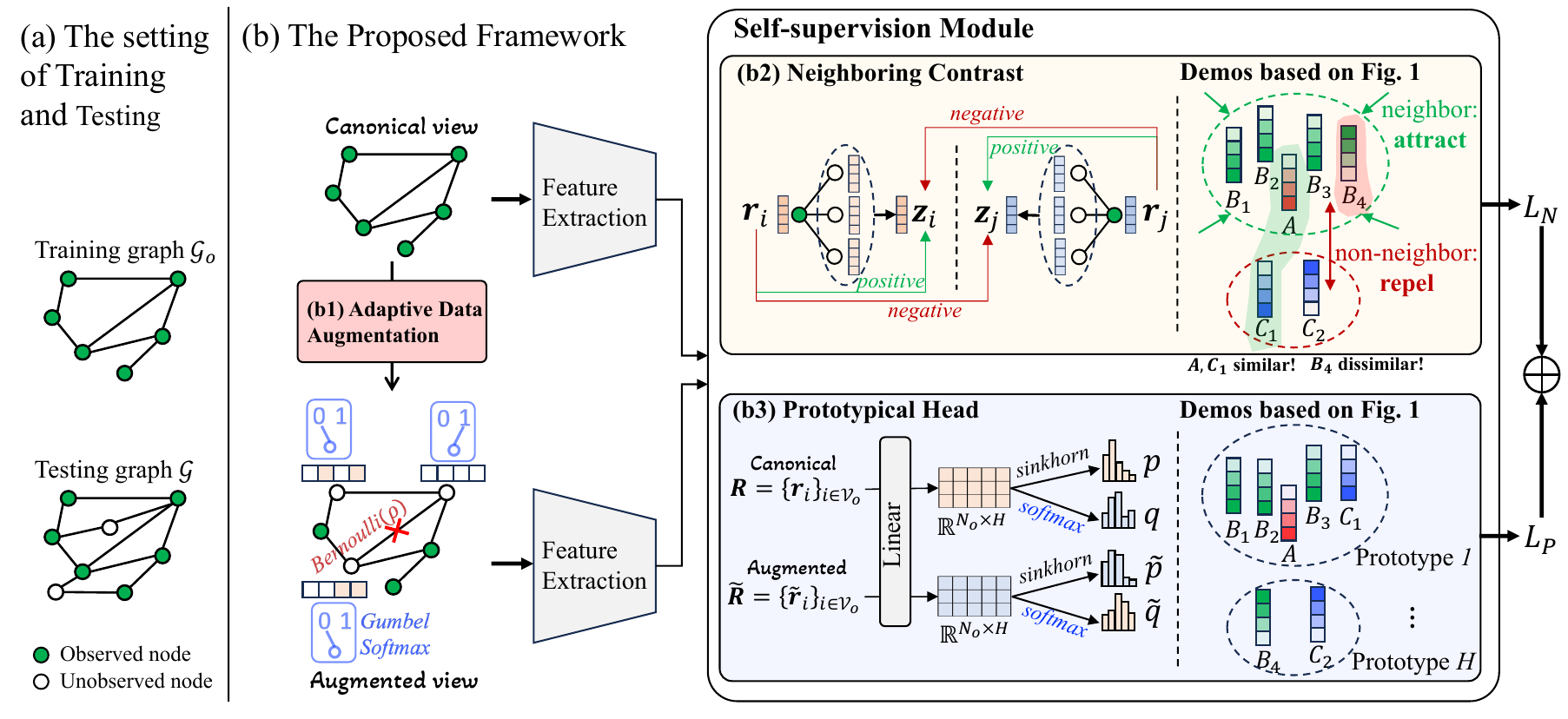}
        \vspace{-6pt}
    \caption{The framework of the proposed KCP}%: (a) The illustrations of training and testing graphs. For inductive Kriging, the unobserved nodes are unavailable in the training stage. (b) the overview of the proposed KCP, which includes three dedicated modules: (1) adaptive data augmentation, (2) neighboring contrast, and (3) prototypical head based on the graph feature extraction module.}
    \label{fig:model}
        \vspace{-15pt}
\end{figure*}
\vspace{-4pt}
\subsection{Problem Definition}
\vspace{-4pt}
%\oldrevise{(1) we could give the dimension of the variables, e.g., $\mathbf{A} \in \mathbb{R}^{N_o \times N_o}, \mathbf{X} \in \mathbb{R}^{N_o \times I}?, \mathbf{Y} \in \mathbb{R}^{(N_o + N_u) \times I}, f(\cdot): \mathbb{R}^{N_o \times I} \mapsto \mathbb{R}^{(N_o + N_u) \times I}?$ (2) We can also define our representation learning task:}

% \oldrevise{here we need to emphasize our problem setting: we only knows the unobserved nodes' topological information, and we don't know their attribute. and justify this is more reasonable in practice}
%Rather than a regular two-dimensional grid, 
In the Kriging task, different locations in the studied region can be formulated as a potential graph structure $\mathcal{G}=(\mathcal{V},\mathcal{E})$, where $\mathcal{V}$ and $\mathcal{E}$ are the nodes and edges sets, respectively. Each geo-location is treated as a node $v\in \mathcal{V}$ and the connections are reflected by the graph adjacent matrix $\mathbf{A}\in \mathbb{R}^{|\mathcal{V}|\times|\mathcal{V}|}$. Thus, it can also be viewed as a graph super-resolution problem: 
Given limited observed nodes $\mathcal{V}_o$ attributes $\mathbf{X}$ ($\mathbf{x}_i$ for node $v_i$'s attributes), the attributes $\mathbf{y}_j$ of an unobserved node $v_j\in \mathcal{V}_u$ can be inferred from the spatial/temporal correlations across nodes. 
{To perform the inductive Kriging, the topological information of unobserved nodes is unavailable in the training stage. 
As shown in Fig.} \ref{fig:model}(a), training is conducted on the subgraph $\mathcal{G}_o\subseteq \mathcal{G}$ composed of $N_o$ observed nodes (with adjacency $\mathbf{A}_o$). In the testing stage,
the un-observations serve as the newly added nodes, and there is $\mathbf{A}\in \mathbb{R}^{(N_o+N_u)\times(N_o+N_u)}$, where $N_u$ denote the number of unobserved nodes and $N_o+N_u=|\mathcal{V}|$. 
In spatiotemporal Kriging, node attributes are usually represented by time series with horizon $T$, i.e., $\mathbf{x}_i=[x^{t+1}_i,\dots,x^{t+T}_i]$. The objective is to find a dedicated function $\mathcal{F}$ which can estimate the attributes of $N_u$ unknown nodes $\hat{\mathbf{Y}}\in \mathbb{R}^{N_u\times T}$ ($\mathbf{Y}$ for ground truth) with the available nodes' information $\mathbf{X}\in\mathbb{R}^{N_o\times T}$ and adjacent matrix $\mathbf{A}$. That is $\mathbf{Z}=\mathcal{F}(\mathbf{X}; \mathbf{A}_o)$ (training) then 
    $\hat{\mathbf{Y}}=\mathcal{F}(\mathbf{Z}; \mathbf{A})$ (testing).
% To address the problem by the SSL, we conduct a two-step operation to explicitly distill the representations: First, the representations of observations are generated by an encoder $\mathbf{R}=\mathcal{F}_{enc}(\mathbf{X}; \mathbf{A}_o)$, where $\mathbf{A}_o\in \mathbb{R}^{N_o\times N_o}$. Then, we recover the Kriging results from the representations by a downstream decoder $\hat{\mathbf{Y}}=\mathcal{F}_{dec}(\mathbf{R}; \mathbf{A})$. 
\vspace{-20pt}\subsection{The architecture of KCP}\vspace{-4pt}
We elaborate on the details of KCP, with the overall framework in Fig. \ref{fig:model}.
Our model mainly contains three tailored components based on a graph feature extraction module: adaptive data augmentation, neighboring contrast, and prototypical head.
The first (Fig. \ref{fig:model}(b1)) is dedicated to performing attribute-level and topological augmentations from the canonical view, and then 
provides the augmented view for feature extraction. 
% \oldrevise{The function of the two needs to be updated}
The last two components achieve self-supervision according to the extracted representations' consistency between the canonical and augmented views:
The former (Fig. \ref{fig:model}(b2)) conducts the contrastive learning by exploiting neighbor and non-neighbor information.
Considering that not all neighbors are definitely similar, and non-neighbors are not always incompatible (as we argued in Fig. \ref{fig:embedding}), the latter (Fig. \ref{fig:model}(b3)) aims to identify similar representations regardless of the adjacency via exchanged prediction, thus refining the positive and recycling the negative from the neighboring contrast.

\paragraph{3.2.1 Graph feature extraction module}
To encode the node representation, we adopt a GNN backbone with two graph aggregation layers (following GraphSAGE \citep{hamilton2017inductive}) for message passing. The rationale lies in 1) it is inductive and thus can be easily applied to new-coming nodes even unseen scenarios;
2) it can model the node-attribute projection (temporal patterns in the Kriging task) and the spatial aggregation across nodes by learnable weights and adjacent matrix $\mathbf{A}$, respectively. For each layer, with the attribute $\mathbf{x}_i$ from $v_i$, the output $\mathbf{x}_i^{\prime}$ after aggregation:
\begin{equation}\small
\mathbf{x}_i^{\prime}=ReLU\left(\mathbf{W}\cdot \left[\mathbf{x}_i, \mathrm{Agg}_{j\in \{j\in \mathcal{V}| \mathbf{A}_{ji}>0\}} \left(\mathbf{W}^t\mathbf{x}_j+\mathbf{b}\right)\right]\right),
\end{equation}
where $ReLU(\cdot)$ is an activation function, $[,]$ represents concatenation, and $\mathrm{Agg(\cdot)}$ means the aggregation function (we simply use $\mathrm{mean}$ in the model). $\mathbf{W}$, $\mathbf{W}^t$, and $\mathbf{b}$ conduct the projection which can simply model the temporal dependencies for the node attributes. After the feature extraction module, the output representation $\mathbf{r}\in\mathbb{R}^{E}$ for each node in two views is yielded, where $E$ is the dimension of the representation, and there are ${\mathbf{R}}\in\mathbb{R}^{N_o\times E}$ for all the nodes in the canonical view and $\tilde{\mathbf{R}}\in\mathbb{R}^{N_o\times E}$ in the augmented view.
\paragraph{3.2.2 Adaptive data augmentation} Data augmentation is a key mechanism for SSL (especially contrastive learning). For graph SSL, there are two families of techniques based on node attribute and topology. However, as argued in \citep{zhu2021graph}, simple data augmentation in attribute/topology domains may not generate diverse contexts. {This issue is more prominent in the Kriging since there are no inherent clues in unknown nodes, and the estimations are completely dependent on contextual information.}
Additionally, noise, such as random missing caused by sensor breakdown in observed nodes, is also notable, which complicates the learning of node representations.

Given the issues, in the training stage, we devise an adaptive data augmentation strategy to corrupt the canonical view, which is illustrated in Fig. S1 in Suppl. % \oldrevise{I think better to have another figure 3.(b) to demo the augmentation}
Specifically, among the observed nodes, we first randomly select $N$ nodes while keeping the remaining nodes unchanged. Then, attribute-level and topological augmentations are sequentially applied to each selected node (More details in Fig. S1 in Suppl. C.1).

% \oldrevise{we need to justify why a node has to adaptively choose from the two options? one is to mimic missing, one is to mimic Kriging?} 
{\textbf{(1) At the attribute-level augmentation}, we allow each sampled node to pick one option between two strategies: 1) feature mask, some attributes on a sampled node are masked under a pre-specified ratio $r_m$, which aims to analog the noise that appears in the known nodes (mimic random missing). We believe it can force the model to focus on the temporal dependencies in each node and thus benefit the robustness of the model; 2) node mask, the attributes of a sampled node are all filled with zeros.} This can be regarded as a special case of feature mask where $r_m=1$, which acts as the unknown nodes (mimic Kriging).

Since it is hard to specify which one option should be picked for each sampled node, inspired by \citet{zhu2021graph}, we design a learning-based operator to achieve the specification adaptively. It can be treated as a binary classification (implemented by a 3-layer multilayer perceptron (MLP)) with the node feature $\mathbf{x}_i$ as input and the binary logit ${\pi}^i$ as output: ${\pi}^i = \mathrm{MLP}(\mathbf{x}_i)=[{\pi}^i_0, {\pi}^i_1]$. Considering that the specification is non-differentiable, we introduce the Gumbel Softmax \citep{jangcategorical} trick as the differentiable approximation. 
During the forward propagation: % \oldrevise{why $y$'s subscript is $i$ and the rest subscript is $j$, $i, j$ mean different attribute or node here?}
\begin{equation}\small
out_i={\arg \max}_m\left(\log \left({\pi}^i_m\right)+g_m\right), m\in \{0,1\},
\end{equation}
where $out_i$ is the output specification for node $v_i$. $out_i=0$ means feature mask and $out_i=1$ denotes node mask. $g_m\sim \mathrm{Gumbel(0,1)}$ is a noise term drawn from the standard Gumbel distribution.
The backward propagation {($\mathrm{MLP}$ update)} is conducted with the temperature parameter $\tau$ by taking the derivative of:
\begin{equation}\small\label{eq:tau}
out_i=\frac{\exp \left(\left(\log \left(\pi^i_m\right)+g_m\right) / \tau\right)}{\sum_{n=0}^1 \exp \left(\left(\log \left(\pi^i_n\right)+g_n\right) / \tau\right)}.
\end{equation}
% where , and we set $\tau=0.5$ in this paper. %\oldrevise{$y_i$ is from 0 to 1, so it is the probability of choosing feature mask or node mask? 0 is 100\% feature mask and 1 is 100\% node mask? the ``Gumbel Softmax'' is the Eq. (4), and it is called Gumbel just because $g_i$ is Gumbel noise, right? yes.}

\textbf{(2) In topological augmentation}, edge drop is performed. \re{Different from traditional random edge dropping, we only drop} the edges connected to a high-centrality node with a probability $\rho$. The underlying prior is that the edge missing does not alter the neighboring information aggregation. 
Based on the centrality, the edges around a sampled node $v_i$ will be dropped according to Bernoulli distribution, i.e., $\mathrm{Bernoulli}(\rho_i), \rho_i=\max((\mathrm{D}_{ii}-{d}_\mathrm{avg})/d_\mathrm{max},0)$, where $\mathrm{D}_{ii}$ is the $i, i$-th entry for degree matrix $\mathrm{D}$, ${d}_\mathrm{avg}$ is the average degree across all the nodes and $d_\mathrm{max}$ is the maximum. This setting ensures that the plain nodes are ignored while the edges around critical nodes are augmented: \re{for the node with a few edges, e.g., only one edge, dropping the only edge will cause the node to have no neighbor, thus during the training, no reference may cause bad inference; in contrast, the central nodes have enough edges, dropping its edges will encourage robust inference from different subsets of neighbors.} 

% Following the G-mix up \cite{han2022g}, the topological augmentation can also be thought of as a special case for graphons mix up: $W_{\mathcal{I}} = (1 - P) \odot W_H$ where the $i, j$-th entry of weighted matrix $P$ is sampled from $\mathrm{Bernoulli}(p_i)$. This setting ensures that the plain nodes below the mediocrity are ignored, while the edges around critical nodes are augmented. Here we also provide its theoretical justification that the augmentation still retains the benefits like G-mix up, 
\textbf{Theorem 3.1} 
\textit{Given the canonical graph $\mathcal{G}$ and augmented graph, 
the homomorphism density $t$ (definations referred to \citep{han2022g}) of the augmented graphon $W'$ is determined by $\Phi$, and $W' = (\mathbbm{1} - \Phi) \odot W$, where the $i, j$-th entry of weighted matrix $\Phi$ is sampled from $\mathrm{Bernoulli}(\rho_i)$. The difference in the homomorphism densities of the canonical graphons $W$ and augmented graphons $W'$  is still upper bounded by }
% \begin{equation}
% t(F, W)=\int_{[0,1] V(F)} \prod_{i, j \in E(F)} W\left(x_i, x_j\right) \prod_{i \in V(F)} d x_i
% \end{equation}
\begin{equation}
\begin{aligned}
\left|t\left(\mathcal{G}, W'\right)-t\left({\mathcal{G}}, W\right)\right| & \leq(1-\lambda) \mathrm{e}\left({\mathcal{G}}\right)\|W\|_{\square}
\end{aligned}
\end{equation}
\textit{where $\lambda=\Pi_{i,j=1}^{N}(1-\Phi_{ij})$, $\mathrm{e}\left(\mathcal{G}\right)$ is the number of the edges in $\mathcal{G}$, and $\|W\|_{\square}$ means cut norm\citep{lovasz2012large}.}

\begin{table*}[t]
%\small
  \caption{Performance comparison. For deep models, the results are obtained through three independent executions and in format \textit{mean$\pm$std}. The best results are in \textbf{bold} and the second-best are \underline{underlined}.
  % \oldrevise{We can add two more rows showing it's statistical/deep learning, transductive/inductive}
  }
  \vspace{-8pt}
  \label{tab:performance}
  \centering
  {
  \resizebox{0.99\textwidth}{!}{%
\begin{tabular}{c|c|cccc|cccc}
    \toprule
    \multirow{2}{*}{Dataset} & \multicolumn{1}{c|}{\multirow{2}{*}{Metric($\downarrow$)}} & \multicolumn{4}{c|}{\textbf{Statistical models}} & \multicolumn{4}{c}{\textbf{Deep models}}\\\cline{3-10}
                                          &                                                        & KNN-IDW & XGBoost & OKriging   & GPMF & BGRL & IGNNK & KCN & KCP(ours)  \\ \hline\hline
\multicolumn{1}{c|}{\multirow{3}{*}{PeMS}}       & \multicolumn{1}{c|}{MAE($\pm$std)}    & 50.95                       & 61.52 & 49.03 & 54.38           & 45.09$\pm$0.311 & 43.98$\pm$0.233             & \underline{43.59}$\pm$0.101            & \textbf{42.29}$\pm$0.122 \\ \cline{2-10} 
\multicolumn{1}{c|}{}                            & \multicolumn{1}{c|}{RMSE }   & 76.33                       & 77.00 & \textbf{64.86} & 70.12   & 70.94$\pm$0.255       & {67.11}$\pm$0.145            & 68.14$\pm$0.177           & \underline{65.37}$\pm$0.154 \\ \cline{2-10} 
\multicolumn{1}{c|}{}                            & \multicolumn{1}{c|}{MAPE } & 36.5\%                         & 42.3\% & 50.1\% & 37.4\%         & 31.9\%   & 32.6\%              & \underline{31.6\%}           & \textbf{29.7\%}  \\ \hline
\multicolumn{1}{c|}{\multirow{3}{*}{NREL}}       & \multicolumn{1}{c|}{MAE }    & 1.778                    & 1.878   & 1.977 & 1.645       & 1.702$\pm$0.019    & 1.813$\pm$0.012             & \underline{1.571}$\pm$0.015           & \textbf{1.569}$\pm$0.013 \\ \cline{2-10} 
\multicolumn{1}{c|}{}                            & \multicolumn{1}{c|}{RMSE  }  & 2.713                    & 3.261   & 2.980 & 2.451    & 2.584$\pm$0.018       & 2.588$\pm$0.017             & \underline{2.366}$\pm$0.011           & \textbf{2.353}$\pm$0.015 \\ \cline{2-10} \cline{2-10} 
\multicolumn{1}{c|}{}                            & \multicolumn{1}{c|}{MAPE } & \underline{33.2\%}                     & 44.7\%    & 45.8\%  & 42.7\%     & 73.8\%       & 36.4\%              & 33.4\%             & \textbf{32.2\%}  \\ \hline
\multicolumn{1}{c|}{\multirow{3}{*}{ Wind}} & \multicolumn{1}{c|}{MAE }    & 1.657                    & \underline{1.562}  & {2.107} & {1.566}    & {1.564}$\pm$0.115       & 1.760$\pm$0.013             & 1.568$\pm$0.022            & \textbf{1.556}$\pm$0.018 \\ \cline{2-10}\cline{2-10}  
\multicolumn{1}{c|}{}                            & \multicolumn{1}{c|}{RMSE }   & 2.339                    & {2.139}   & 2.789 & 2.154   & \textbf{2.049}$\pm$0.016        & 2.489$\pm$0.017             & 2.123$\pm$0.018            & \underline{2.096}$\pm$0.012 \\ \cline{2-10} \cline{2-10} 
\multicolumn{1}{c|}{}                            & \multicolumn{1}{c|}{MAPE } & 39.1\%                     & 32.3\%    & 55.3\%  & 37.6\%  & \textbf{30.3\%}  & {55.1\%}              & {33.4\%}            & \underline{31.6\%} \\
    \bottomrule
  \end{tabular}}}
      \vspace{-12pt}
\end{table*}

Theorem 3.1 suggests that topological augmentation is a special case of G-Mixup while sampling entries in $\mathrm{Bernoulli}(\rho)$ and the augmentation still retains the benefits of G-Mixup (i.e., promising generalization and robustness). The detailed proof is in Suppl. C.2.

\paragraph{3.2.3 Neighboring Contrast}
To guide the model to estimate unobserved nodes by encoding neighboring information, we propose to supervise the target nodes with the representations of their neighbors. 
Rather than node-to-node or node-to-graph contrast, we primarily emphasize node-to-neighbor contrast since the Kriging task should not refer to a specified node nor to the whole graph; instead, the neighbors.
For an anchor node $v_i$, its representation $\mathbf{r}_i$ and its aggregated neighboring representations in the other view constitute a positive pair (${\mathbf{r}}_i, \mathbf{z}_i$), while the neighboring aggregation of other nodes is set as the negative pair of $\mathbf{r}_i$.
Since neighbors are not equally important, the aggregation is adopted by an attention-based readout: \\
\begin{equation}\small
\mathbf{z}_i=\mathbf{W_2}\cdot(\sum\nolimits_{j \in N_k(i)} \alpha_j \mathbf{r}_j),
\end{equation}
where 
$\alpha_i = \nicefrac{\exp \left(\mathbf{W_1} \mathbf{r}_i\right)}{{\sum_{j \in N_k(i)} \exp \left(\mathbf{W_1} \mathbf{r}_j\right)}}$. $\mathbf{W_1}, \mathbf{W_2}$ are the trainable matrices, and $N_k(i)$ is a set that contains the top-$k$ nearst neighbors of $v_i$.
With the canonical view and the augmented view, we maximize the agreement between node representations by the noise contrastive estimation loss:
\begin{equation}\small
\begin{aligned}
\mathcal{L}_{N}=& -\frac{1}{|\mathcal{V}|}\sum_{i \in \mathcal{V}}\left[\log \sigma\left(\mathcal{D}\left(\mathbf{r}_i, \mathbf{z}_i\right)\right)\right.\\&+\left.\mathbb{E}_{w \sim \hat{\mathbb{P}}}\left[\log \left(1-\sigma\left(\mathcal{D}\left(\mathbf{r}_i, \mathbf{z}_w\right)\right)\right)\right]\right],
\end{aligned}
\end{equation}
where $\sigma$ is sigmoid function {and $\mathbf{z}_w$ is a negative sample}. $\hat{\mathbb{P}}$ is the distribution of negative samples. $\mathcal{D}$ measures the agreement between two vectors, i.e., their cosine similarity $\mathcal{D}(\mathbf{x},\mathbf{y})=\nicefrac{\mathbf{x}\cdot \mathbf{y}}{|\mathbf{x}||\mathbf{y}|}$. {As a result, in Fig. \ref{fig:model}(b2), target A will be attracted together with its neighbors $B_1$-$B_4$, and repelled away from its non-neighbors $C_1$-$C_2$, even though $C_1$ is similar to $A$, not to $B_4$.}
\paragraph{3.2.4 Prototypical Head}
% \oldrevise{Again, the purpose of prototype head should be adjusted as "refine and recycle" the information from contrast module} 
Although neighboring contrast utilizes neighbor and non-neighbor information to obtain effective representations under the guidance of Kriging's common practice, it is not always rational since non-neighboring nodes may share similarities in certain time series patterns such as change trends, peaks, or slopes. For example, in traffic, similar traffic flow patterns can appear even in distant intersections due to similar functional areas. 
To play a ``refine and recycle'' role for the neighboring contrast component, we further introduce a prototypical head \citep{liu2021self} module for self-supervision. 

Assume there are $H$ typical patterns, called prototypes, among all node attributes that can be clustered. The prototypical head aims to uniformly assign prototype labels to each node with subject to $\sum_i q_{ih}=1$ and $\sum_h q_{ih}=1$, where $q_{ih}$ is the assignment probability with prototype $h$ to node $v_i$. %\oldrevise{The $i, j$ here should mean different ij from Eq.3 and 4?}. 
The supervisory signal comes from an intuitive principle: the representations from two views for an anchor node should yield similar assignment probabilities to the same prototype.
{To this end, we project the representation into a new latent space $\mathbb{R}^H$ by a learnable head $\mathbf{H}\in \mathbb{R}^{E\times H}$.} Specifically, for node $v_i$, there is $\mathbf{c}_i=\mathbf{r}_i\cdot\mathbf{H}=[c_{i1},c_{i2},...c_{iH}]$. The score for assigning prototype $h$ to node $v_i$ can be calculated as $p_{ih}=\nicefrac{\mathrm{exp}(c_{ih})}{\sum_{j}^H\mathrm{exp}(c_{ij})}$.
%\begin{equation}
%    p_{ij}=\frac{\mathrm{exp}(c_{ij})}{\sum_{h}^H\mathrm{exp}(c_{ih})}.
%\end{equation}
Then, the assignment problem can be cast as an optimal transport problem and the optimal assignment probability $q_{ih}$ can be computed as the soft labels by the off-the-shelf iterative Sinkhorn algorithm \citep{cuturi2013sinkhorn} under the same input $\mathbf{c}_i$, {i.e., $q_{ih}=\mathrm{Sinkhorn}(c_{ih})$}. % \oldrevise{is there any equations to calculate q? at least we need to specify what is the input to calculate q?}. 
% {to response: The initialization of q in line 234 is not provided.}
For the prototype vectors from the canonical view and augmented view (denoted by $\mathbf{c}_i$ and $ \tilde{\mathbf{c}}_i$, respectively) of node $v_i$, we construct the supervision by cross-predicting the pair-wise loss from two views: using the assignment probability from augmented view $\tilde{q}$ to guide the score ${p}$ from the canonical view and vice versa:
\vspace{-6pt}
\begin{equation}\label{eq:proto}
\small
\begin{aligned}
    \ell\left(\mathbf{c}_i, \tilde{\mathbf{c}}_i\right)&=-\sum_h^H \left[\tilde{q}_{ih} \log p_{ih}+q_{ih} \log \tilde{p}_{ih}\right] \\&=-\sum_h^H \left[\tilde{q}_{ih} \log \frac{\exp \left({c}_{ih}\right)}{\sum_j \exp \left({c}_{ih}\right)}+{q}_{ih} \log \frac{\exp \left(\tilde{c}_{ih}\right)}{\sum_j \exp \left(\tilde{c}_{ih}\right)}\right].
\end{aligned}
\end{equation}
When migrating all the nodes, the total self-supervised loss for the prototypical head module is:
\begin{equation}\small
\mathcal{L}_{P}= \frac{1}{|\mathcal{V}_o|}\sum_{i\in \mathcal{V}_o}\ell\left(\mathbf{c}_i, \tilde{\mathbf{c}}_i\right).
\end{equation}

As a result, shown in Fig. \ref{fig:model}(b3), only the similar embeddings for target $A$ are selected into the same prototype, i.e., nodes $B_1$-$B_3$ and $C_1$, for $A$'s inference.

The final training loss is $\mathcal{L}_{SSL}=\mathcal{L}_{N}+\mathcal{L}_{P}$. \zs{We also summarize the computational complexity of the KCP, which is $\mathcal{O}(\sum^{L-1}_{i=1}NE_iE_{i+1}+\sum_{i=1}^{L_n}(N+K)E_i+\eta HN+NEH)$ (details in Suppl. B).}

\section{Experiments and analysis}
% In this section, we first set up the experiments, and then provide the performance comparison and result analysis.
\vspace{-4pt}
\subsection{Experiments setup}
\vspace{-4pt}
\textbf{Dataset}
For a broader performance evaluation of the proposed model, we conduct Kriging experiments on three publicly available time series datasets, \zs{since they are representative in particular application domains and are all with geographical properties. They are
1) \textbf{PeMS}: The dataset aggregates a 5-minute traffic flow across 325 stations, which is collected from the Caltrans Performance Measurement System over 2 months (January 1st, 2017 to March 1st, 2017).  2) \textbf{NREL}: It records 5-minute solar power output from 137 photovoltaic plants in Alabama in 2006, which is extracted from \citep{wu2021inductive}. 3) \textbf{Wind}: This dataset records onshore renewable energy generation for Greece, which contains hourly wind speed aggregation on 18 installations over 4 years \citep{vartholomaios2021short}.}

%, which are described in Table \ref{tab:data_desc}. 
%\oldrevise{This paragraph or Table 1, only one of them is enough. Maybe only table 1 is already enough and this paragraph for appendix} 
%\renewcommand\arraystretch{1.2}
In each dataset, $80\%$ randomly selected nodes are set as observations (i.e., $\mathcal{V}_o$) for training, and the remaining are regarded as unobserved nodes $v_u\in\mathcal{V}_u$ for testing. The data are normalized by the min-max scaler. More details are in Suppl. C.3. \\
\textbf{Training and testing}
For a full graph $\mathcal{G}$, we only sample a subgraph $\mathcal{G}_o$ with $|\mathcal{V}_o|$ nodes for training. The Kriging task is conducted under a pretraining and then fine-tuning paradigm. 
During pretraining, we optimize the proposed SSL loss %\oldrevise{the two losses are equally important? yes} 
for estimating the representations of augmented nodes in the augmented view. When fine-tuning, we use a 3-layer MLP to recover the node attributes according to the estimated representations of target nodes and finetune the parameters of the feature extraction module under the mean absolute error loss.
In the testing stage, the $|\mathcal{V}_o|$ nodes are treated as known nodes while the remaining nodes in $\mathcal{G}$ serve as new-coming and unobserved locations, which are unavailable at the training stage.
% \begin{figure}[h]
%     \centering
%     \includegraphics[width=0.6\columnwidth]{figs/data_split.pdf}
%     \caption{Data split. \oldrevise{can be put together with what I meant in Fig.2b}}
%     \label{fig:datasplit}
% \end{figure}
\\
\oldrevise{\textbf{Baselines} We choose the following benchmarks: (1) statistical models: KNN with inverse distance weights, \textbf{KNN-IDW} for short, \textbf{XGBoost} \citep{chen2016xgboost}, \textbf{OKriging} \citep{bostan2017basic}, \textbf{GPMF} \citep{strahl2020scalable}, and (2) GNN-based models: \textbf{BGRL} \citep{thakoor2021large}, \textbf{IGNNK} \citep{wu2021inductive}, and \textbf{KCN} \citep{appleby2020kriging}. The details about the methods are in Suppl. C.4.}
%\textbf{Baselines} \oldrevise{This part could be put in the appendix} We compare six advanced Kriging models involving classical machine learning and deep learning methods. They are 
%1) KNN-IDW: A KNN model which incorporates inverse distance weighted interpolation for Kriging. The weighted average of the $K$ nearest observed nodes to the unobserved node is taken as the estimation, and the weight is associated with the entry in the adjacent matrix. We set $K=5$ in this paper.
%2) XGBoost \citep{chen2016xgboost}: We train an extreme gradient boosting model, in which the geolocations are regarded as the input, and the values of the attributes are the output.
%3) OKriging \citep{bostan2017basic}: Ordinary Kriging. It estimates the unknown nodes with a known variogram under the Gaussian process, which is a typical spatial interpolation model.
%4) GPMF \citep{strahl2020scalable}: It is a graph-based prior probabilistic matrix factorization, in which the graph structure is used as the side information to achieve the matrix factorization.
%5) IGNNK \citep{wu2021inductive}: It constructs dynamic subgraphs by random sampling and uses diffusion graph convolutional network \citep{li2018dcrnn_traffic} for the spatiotemporal Kriging.
%6) KCN \citep{appleby2020kriging}: A graph convolutional network for Kriging, which makes direct use of $K$ nearest neighboring observations for graph message passing. We also set $K=5$.

%For the neural network models, we split the time series attributes in each node by slide window with size $SW=24$ for input.
\vspace{-4pt}
\subsection{Results and detailed analysis}
\vspace{-4pt}
%\oldrevise{we could add the figure: predicted target attribute and the neighbors' attribute it use to construct, to show two case: (1) neighbors' attribute could be misleading (2) non-neighbors' attribute could also help}

% To evaluate our KCP, 
We conduct extensive experiments by answering the following questions: \textbf{Q1}: 1) Does the proposed model outperform the state-of-the-art baselines? 2) How will the models perform with a higher percentage of unseen nodes? \textbf{Q2}: What are the distinctive/advantages of the learned representations by SSL for Kriging? \textbf{Q3}: Which component is the most important in the proposed model? \textbf{Q4}: Any cases to support: misleading neighbors and constructive non-neighbors?

\vspace{-1pt}\textbf{Kriging task performance (Q1.1):} 
In Table \ref{tab:performance}, we summarize the performance of the proposed KCP and baselines on the three datasets. The best results are in bold and the second-best ones are underlined.
Three commonly used metrics are adopted to evaluate the Kriging models, \oldrevise{i.e., MAE, RMSE, and MAPE (definitions in Suppl. C.5)}. The lower metrics indicate better performance. %which are 1) mean absolute error (MAE), 2) root mean square error (RMSE), and 3) mean absolute percentage error (MAPE). The lower metrics indicate better performance.

It can be seen that the KCP outperforms its peers across almost all the datasets (about a maximum of $ +6\%$ improvements than the second-best one), achieving the lowest metrics except for the RMSE in the PeMS dataset. 
Specifically, compared to statistical models, GNN-based deep models typically obtain better Kriging performance.
KNN-IDW, the straightforward spatial interpolation, is surprisingly competitive in Kriging, since it goes beyond simple neighbor averaging and incorporates inverse distance-weighted interpolation which facilitates the neighboring information aggregation.
XGBoost also shows promising performance in the Wind dataset. 
The reason may be that it takes latitude and longitude as input, which is more sensitive to geographic location, whereas wind speed often exhibits non-skewed spatial distributions and relates to geolocation.
The BGRL achieves moderate outcomes, as there is no specific SSL design for the Kriging task.
KCN outperforms the baselines on the PeMS and NREL datasets, while IGNNK is also impressive under RMSE. Additionally, a representative scheme of matrix factorization, GPMF, also achieves good performance. %It works without the training set, which is only trained by the known nodes in the test set and then estimates the unknown nodes. 
However, it lacks inductive capability, thus cannot handle the new nodes without re-training.

\vspace{-1pt}
\textbf{Different observation ratio (Q1.2):} 
To evaluate the model's performance under different ratio observations, we look through the Kriging results by varying the unsampled ratio of unknown nodes. The unobserved ratio ranges from $20\%$, $50\%$, and $70\%$. For clarity, $20\%$  means to infer $20\%$ unobserved nodes by $80\%$ known nodes. %\oldrevise{it should be opposite?: In Table 3, $20\%$ has the best performance, meaning 20\% is the easiest = least unobserved (most observed): so it is 20\%  unobserved, 80\% observed} 
The results of the advanced GNN-based baselines and our model are summarized in Table \ref{tab:kriging_ratio}. It can be seen that the superiority of the proposed model still holds with the changes in the unobserved ratio, which demonstrates its powerful Kriging ability across various scenarios.%, achieving about 2.8\%$\sim$5.2\% improvements.
\begin{table}[t]
		\centering
          \caption{The Kriging results on PeMS across different unsampled proportions}
          \vspace{-6pt}
        \label{tab:kriging_ratio}
    \resizebox{1\columnwidth}{!}{
    % {
\begin{tabular}{cccc|ccc}
\toprule
\multicolumn{1}{c}{\multirow{2}{*}{Models}} & \multicolumn{3}{c}{\textbf{MAE}}                 & \multicolumn{3}{c}{\textbf{RMSE}} \\\cline{2-7}
\multicolumn{1}{c}{}                        & 20\%           & 50\%           & 70\%           & 20\% & 50\%  & 70\%       \\\hline
IGNNK                                       & 43.98          & 68.21          &  110.93           & 67.11     & 93.07     &  163.60       \\\hline
KCN                                         & 43.59          & 62.19          & 88.53          & 68.14     & 93.61     & 122.60        \\\hline
{KCP }                         & \textbf{42.29} & \textbf{59.50} & \textbf{87.74} & \textbf{65.37}     & \textbf{86.49}     & \textbf{114.53}     \\\hline
{\textbf{Imp.} ($\uparrow$)}                         & {3.0\%} & {4.5\%} & {0.9\%} & 3.1\%    & 4.3\%     & 6.7\%    \\
\bottomrule
\end{tabular}}
\end{table}

\begin{figure}[t]
    \centering
    \includegraphics[width=1\columnwidth]{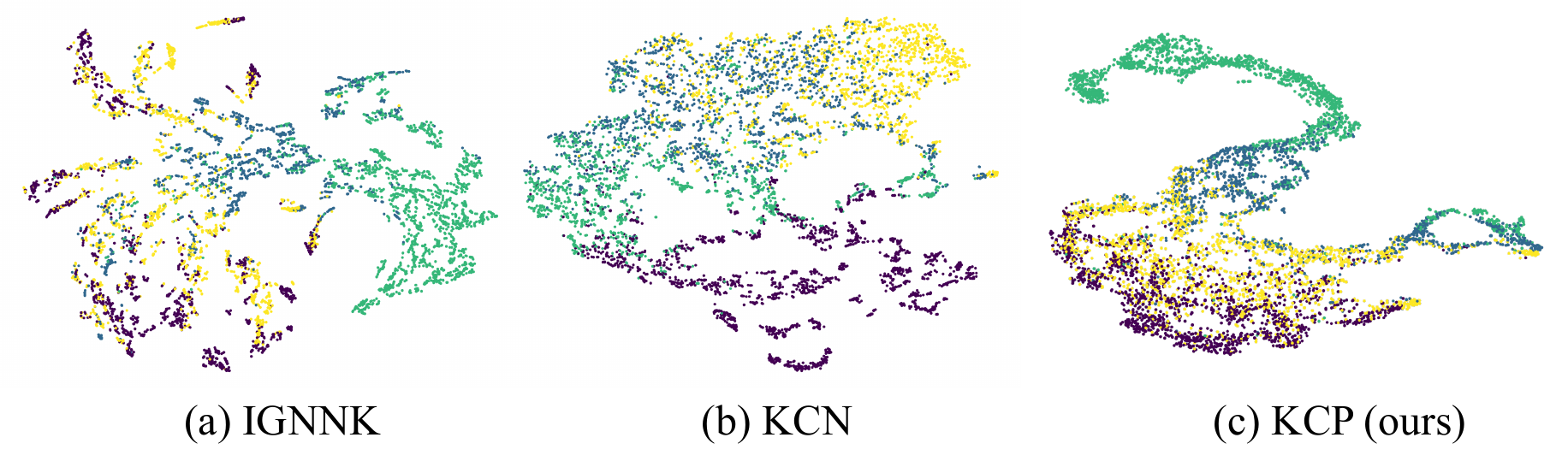}
    \caption{Embeddings of learned representations on PeMS. According to the ground truth, similar time series are clustered into the same class (same color). }
    \label{fig:emb_comp}
        \vspace{-12pt}
    \end{figure}
\begin{figure}[t]
		\centering
            \includegraphics[width=\columnwidth]{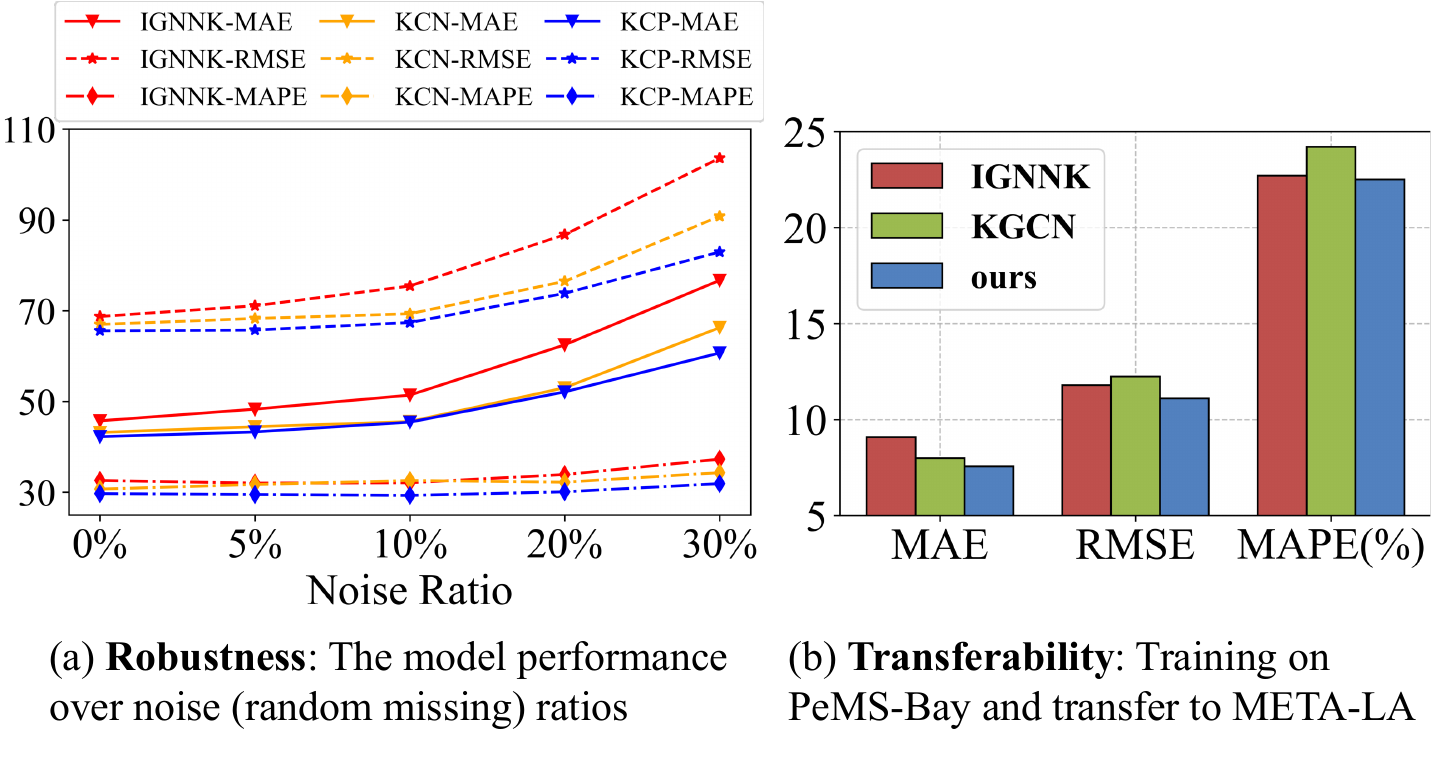}
            \caption{(a) Robustness and (b) Transferability.}
            \label{fig:maskratio}
            \vspace{-14pt}
\end{figure}

\textbf{The learned representations (Q2):} 
Next, we visualize the learned representations of the time series across all the unobserved nodes. To identify inherent patterns, KMeans is adopted to cluster them according to the ground truth, making similar data appears in the same class. %Then, we assign the distinct class ID to the corresponding learned representations as labels. 
We use t-SNE to visualize the learned representations of the unobserved time series in Fig. \ref{fig:emb_comp}. %, whose dimension is reduced by t-SNE. Different colors mean distinct patterns (labels) of time series. 
It can be found that our representations are more compact, and different classes are with better separability in the latent space than IGNNK and  KCN. \textbf{\textit{This proves that our KCP learns the representation with promising quality.}}

% \begin{wrapfigure}{r}{0.38\textwidth}
% 
%     \centering
%     \includegraphics[width=0.38\textwidth]{figs/maskratio.png}
%     \caption{The performance of models across different noise ratios.}
%     \label{fig:maskratio}
%     % \vspace{-20pt}
% \end{wrapfigure}
% \vspace{-12pt}
\textbf{Robustness to confronting noise (Q2):} 
In practical situations, noise is a common problem. For example, in the traffic monitoring system, the records in some monitored intersections may be missing when the devices are temporarily offline caused by unexpected events such as high temperature and network error. Therefore, the robustness to noise is also an important factor to be accounted for in the Kriging model.
To verify the robustness of the models against noise, in the inference stage on PeMS, we set 5\%, 10\%, 20\%, and 30\% random missing for the attributes in the known nodes and investigate the estimations of the unknown nodes. The results are shown in Fig. \ref{fig:maskratio}(a). With the noise ratio increasing, the performance of all the models is degraded, while our model can still beat the strong GNN-based baselines. As the noise ratio increases, its superiority becomes more pronounced. \textbf{\textit{This proves the robustness of the representations from KCP.}}

% \hfill
\begin{figure*}[t]
    \centering
    \includegraphics[width=0.99\textwidth]{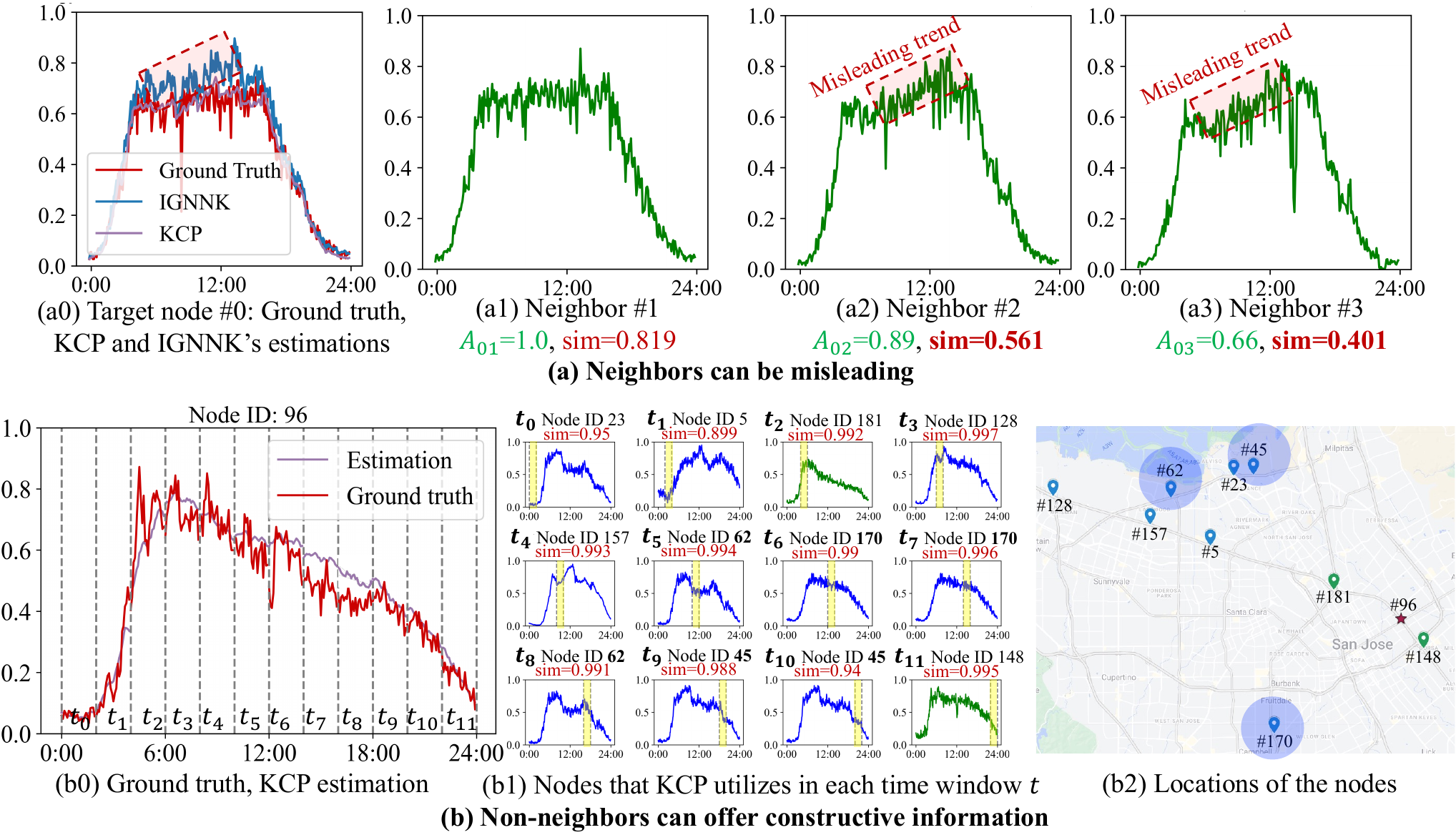}
    \caption{\textbf{(a) Neighbors can be misleading:} Baseline (IGNNK) uses misleading patterns from neighbors \#2 and \#3, who are close but less similar; \textbf{(b) Non-neighbors can be constructive}: KCP dynamically utilizes the patterns from the most similar nodes at each time $t$, \re{e.g., at $t_0$, \#23 is used since similarity = 0.95}: some are neighbors (green), and some are non-neighbors (blue). Non-neighbors \#45, \#62, \#170 used twice.}%: (a) The illustrations of training and testing graphs. For inductive Kriging, the unobserved nodes are unavailable in the training stage. (b) the overview of the proposed KCP, which includes three dedicated modules: (1) adaptive data augmentation, (2) neighboring contrast, and (3) prototypical head based on the graph feature extraction module.}
    \label{fig:neighbors-visual}
        \vspace{-12pt}
\end{figure*}

\begin{figure}[t]
\centering
\includegraphics[width=0.99\columnwidth]{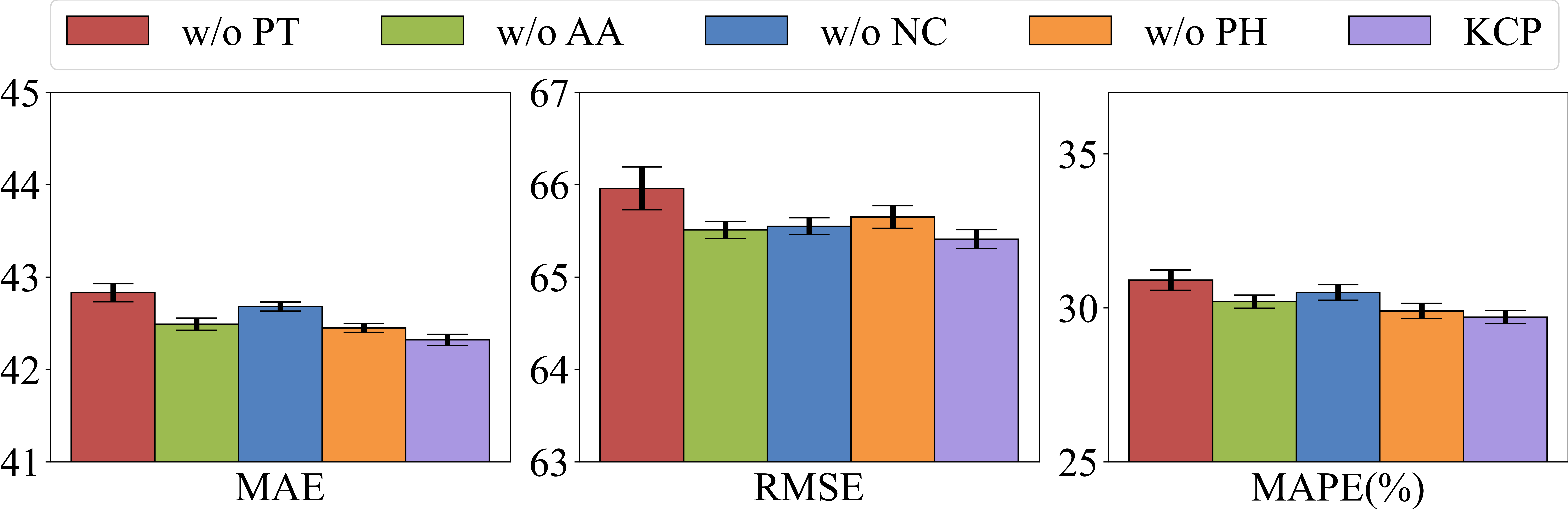}
\vspace{-10pt}
    \caption{Ablation results on PeMS}
    \label{fig:ablation}
    \vspace{-20pt}
\end{figure}

% \begin{wrapfigure}{r}{0.3\textwidth}
%     \centering
%     \includegraphics[width=0.28\textwidth]{figs/transfer.png}
%     \caption{Transferability results}
%     \label{fig:transfer}
% \end{wrapfigure}

\textbf{Transferability validation (Q2):} 
To evaluate the inductive performance with the unseen scenario, we exploit two independent datasets with the same attributes for evaluation, i.e., PeMS-Bay and METR-LA. 
Both two datasets are extracted from \citep{li2018dcrnn_traffic} with 5-min traffic speed aggregation.
The model is trained on PeMS-Bay while transferring without retraining in METR-LA. The testing results are shown in Fig. \ref{fig:maskratio}(b). The experiments show that the KCP can be generalized to an unseen scenario and achieves better transferability.

\textbf{Ablations study (Q3):} 
Since KCP contains three tailored components for spatiotemporal Kriging, we perform the following ablation experiments to provide insights into these components specifically.
1) without pretraining (w/o PT). The model is trained in an end-to-end way for Kriging without the SSL module.
2) without adaptive augmentation (w/o AA). We replace the adaptive augmentation module with a general view augmentation module, which simply masks all the selected nodes.
3) without the neighboring contrast (w/o NC). We remove the neighboring contrast module.
4) without the prototypical head (w/o PH). The prototypical head module is removed from the KCP. The experimental results on PeMS dataset are shown in Fig. \ref{fig:ablation}. It can be concluded that (1) Compared to the model w/o PT, the introduce of SSL is beneficial to the model performance (about $2\%\sim 3\%$ improvements).
(2) The neighboring information aggregation is most crucial for the Kriging task, since the model without neighboring contrast (w/o NC) component causes the largest degradation. (3) The absence of the prototypical head component (w/o PH) demonstrates worse results than the proposed model, %and even it is inferior to the supervised learning-only model under the MAPE metric ($-5\%$ degradation). This 
which indicates that the refining and recycling operations are still noteworthy. (4) The adaptive augmentation component is effective and helpful to the SSL model since its removal also hurts the model performance partly. We also analyze the parameter sensitivity of each component in Suppl. C.6.
% \begin{figure}[t]
%     \centering
%     \includegraphics[width=0.6\columnwidth]{figs/ablation.png}
%     \caption{Ablation study}
%     \label{fig:ablation}
% \end{figure}

\textbf{Visualization of Two Cases (Q4):} (1) Misleading Neighbors and (2) Constructive Non-Neighbors. Fig. \ref{fig:neighbors-visual}(a) shows: baselines such as IGNNK could get misled by the neighbors, e.g., IGNNK predicts an upward trend in the red box due to the misleading patterns from two quite dissimilar yet close neighbors \#2 and \#3. Fig. \ref{fig:neighbors-visual}(b) shows: Our KCP makes the best of the features from different neighbors and non-neighbors at each time slot $t$.  We also spatially visualize the kriging performance based on PeMS in Suppl. C.7.
\vspace{-10pt}
\section{Conclusion}
\vspace{-10pt}
In this paper, we propose a self-supervised learning model for spatiotemporal Kriging. 
Rather than directly predicting the attributes of unobserved nodes, we achieve more robust Kriging by estimating the representations and then recovering them under the SSL framework.
The tailored adaptive data augmentation and SSL modules can encourage data diversity and facilitate the model fully exploit helpful information, respectively. 
% Specifically, the adaptive augmentation module incorporates data-driven attribute augmentation and centrality-based topology augmentation over the spatiotemporal Kriging graph data, which encourages data diversity and facilitates the estimation. 
We not only enhance the neighboring aggregation ability of the GNN backbone by the neighboring contrast, but also emphasize the importance of refining the neighboring and recycling non-neighboring information by the prototypical head when constructing the SSL module.

% The proposed approach convincingly outperforms baselines which include statistical Kriging models and GNN-based deep methods on diverse types of real-world datasets. The limitation is that the temporal correlations inside the node attributes may not be fully utilized by the GNN backbone, and bringing in sequence models can provide new insights to tackle the~issue.

\bibliography{main}

\newpage

\appendix
\onecolumn
\section{More detailed related work}
% {better to categorize them as: transductive v.s. inductive; statistical methods v.s. deep methods}
\subsection{Kriging and SSL methods}
As mentioned in the main text, the fundamental intuition behind Kriging is to model the spatial correlation across observed points to estimate the attributes at unobserved locations \citep{krige1951statistical,goovaerts1998ordinary}. 
The taxonomy of Kriging methods can be categorized as \textit{transductive} and \textit{inductive}. (1) {Transductive models require all the nodes to be present during training, and it cannot learn the representation for the unseen nodes in a natural way: %the graph in testing not to exceed the scope of the observed graph in training. When dealing with unseen nodes,
re-training the model with the new nodes is needed. Classic models such as matrix factorization, DeepWalk, and GCN are by default transductive.} % to have can only tackle the fixed-size graphs and have already seen the topological structure of unobserved nodes during training, which cannot handle the newly added nodes even unseen scenarios without retraining. 
(2) Inductive models instead can directly handle the new nodes that are unseen during training. %The latter is once trained and for all. 
It can accommodate dynamic graphs and learn the representation of unseen nodes. {Here we will introduce more details in Transductive Kriging.}

\textbf{Transductive Kriging}\label{app_rw} Recently, there has been a significant focus on learning-based variants of Kriging to capture spatiotemporal patterns dynamically. (1) Several previous \textbf{GCN-based studies} construct spatiotemporal affinity matrices about the observed and unobserved nodes, and infer the city-wide traffic volume with the constraints of the spatiotemporal consistency \citep{meng2017city,yu2019citywide,dai2021temporal}. 
However, the predefined affinity matrices usually make the model transductive, and thus they are unable to effectively generalize to infer new nodes. (2) Another transductive stream emphasizes the \textbf{statistical approaches}, matrix/tensor factorization \citep{li2020long}, for filling the unobserved attributes, thus the spatiotemporal data can be decomposed into the product of low-rank matrices \citep{jain2014provable}. In this case, the node features are arranged under matrix/tensor forms, in which the unobserved nodes are embodied as completely-missing rows. %By minimizing the errors between known elements in the corrupted matrix and the reconstruction from the decomposition, the attributes in unseen nodes can be completed incidentally. 
The geographic structure is introduced as auxiliary information or priors \citep{zhou2012kernelized,strahl2020scalable,chen2023bayesian}. To estimate the network-wide traffic volume, \citet{zhang2020network} propose to incorporate floating car data into a geometric matrix completion model and add a divergence-based spatial smoothing index to measure the difficulty of estimation in each road segment.
\citet{lei2022bayesian} present a Bayesian kernelized matrix factorization model to capture the spatiotemporal dependencies among the data rows and columns, which is regularized by Gaussian process \citep{rasmussen2006gaussian} priors over the columns of factorized matrixes.
Such methods are still transductive and lack real-time inference capabilities, since newly added nodes will increase the number of rows in a matrix, and thus re-factorization is inevitable.

% \subsection{Self-supervised learning on graphs}\label{app_wonei}
\textbf{Self-supervised learning on graphs} In SSL, predictions or labels are generated from raw data and guide the model's learning by pretext tasks. Contrastive learning, as a sub-domain of SSL, aims to learn  general representations by %find similar and distinct elements from the augmentations and 
maximizing the mutual information (MI) between positive (similar) and negative (distinct) samples that are generated from data augmentation \citep{ci2022fast,tang2022unifying}. The key lies in how to construct and define the positive/negative samples, which affects the representation quality largely. In the context of Kriging, the choice of positive/negative samples could mean which nodes will be used to infer the unseen node. Graph SSL \citep{you2020does, liu2022graph,lin2023dynamic} %is highly related and 
offers potential solutions. The augmentation of graph SSL is usually either in node attributes or structural topology \citep{hassani2020contrastive,you2020graph,zhu2021graph}.

When specifying positive/negative, similar to the Kriging norm, neighbors are usually considered as positive samples \citep{mao2022jointly} and non-neighbors as negative samples. However, scholars also realized the problem with the negative sample definition. \citet{grill2020bootstrap} and \citet{ thakoor2021large} stated that it is rather difficult to contrast a realistic but semantically dissimilar augmented sample; thus, a contrast loss without negative samples was proposed. \citet{kiryo2017positive,chen2020self,tonekaboni2021unsupervised} raised the concern of \textit{negative sampling bias}, where blindly drawing samples from the distribution outside of the positive samples may result in negative samples that turn out quite similar to the reference; thus, ``Positive-Unlabeled'' learning is proposed: outside of the positive region, %it is not strictly negative; instead, 
it is an unlabeled region with a weighted combination of $w$ positive and $1-w$ negative.

We improve the dilemma of positive/negative even further, considering both positive and negative are not strictly cut off, both with exceptions. We propose ``Contrastive-Prototypical'' learning for Kriging, where the contrastive module coarsely defines positive/negative as neighboring/non-neighboring and the prototypical module refines the positive and recycles the negative. Besides, an adaptive augmentation is also proposed to choose the feature mask and node mask in a probabilistic manner, encouraging higher diversity.

\subsection{Methods that consider non-neighbors} \label{app_non_nbs}

Outside the scope of Kriging, there are methods that also consider the nodes that are not physical neighbors. They mainly introduce more semantic graphs such as Point of Interest (POI) \citep{li2020tensor}, transition \citep{mao2022jointly}, and connectivity \citep{geng2019spatiotemporal} (which are the most popularly defined graphs), besides the topological graph. These semantic graphs are either incorporated as additional graph Laplacian penalties in statistical models such as matrix factorization \citep{yang2021real} and tensor decomposition \citep{li2020tensor}, or constructed as multi-view graphs in GCN-based models \citep{geng2019spatiotemporal}. However, these explicitly defined graphs require domain knowledge, which might not be wholesome to explain why two nodes are similar. Moreover, these methods are transductive. {INCREASE \citep{zheng2023increase} is the first inductive method that explicitly defined the three graphs, i.e., spatial proximity, function similarity, and transition probability for Kriging. It shows that, with more graphs considered, the performance could consistently increase. However, the questions are: are three graphs enough and accurate to describe how different nodes utilize each other's information? If the nodes are correlated with ten different patterns, do ten graphs need to be constructed? How to construct graphs that are outside the scope of human domain knowledge?} Therefore, we propose to select similar or dissimilar nodes and show nodes utilize each others' information in a learning base: {no explicit definitions are required, and most likely, the patterns that are not under the awareness of domain knowledge could also be captured.}

\zs{\section{Computational complexity analysis}
The proposed KCP includes three main modules, each contributing to the overall computational complexity. The complexities of these modules are as follows: \\
(1) GNN (i.e., GraphSAGE) is an off-the-shelf component and its complexity is $\mathcal{O}(\sum^{L-1}_{i=1}NE_iE_{i+1})$, where $L$ is the number of GNN layers, $N$ indicates the number of nodes in the graph, and $E_i$ means the dimension of the $i$-th layer representations (embeddings).
(2) The neighboring contrast module is designed to supervise target nodes using representations from their neighbors, which involves aggregating features from $K$ neighbors for each node in the graph. Consequently, the computational complexity of this module is expressed as $\mathcal{O}(\sum_{i=1}^{L_n}(N+K)E_i)$, where $L_n$ indicates the number of aggregation layers (defaulted to 1) and $K$ participates in the attention-based readout.
(3) The prototype head incorporates a linear projection and the Sinkhorn algorithm. Since the implementation of the Sinkhorn involves alternately iterative normalization of the rows and columns for the matrix, during the model training, it contributes to a computational complexity denoted as $\mathcal{O}(\eta HN)$, where $\eta$ represents the number of iterations and $H$ is the number of prototypes. Associated with the $\mathcal{O}(NEH)$ complexity of the linear projection, the overall computational complexity of the prototype head is expressed as $\mathcal{O}(\eta HN+NEH)$.
To sum up, the total computational complexity of the KCP model is the summarization of the three modules, that is, $\mathcal{O}(\sum^{L-1}_{i=1}NE_iE_{i+1}+\sum_{i=1}^{L_n}(N+K)E_i+\eta HN+NEH)$.}
% \vspace{-10pt}
\section{Implementation Details}
% In this section, we first set up the experiments, and then provide the performance comparison and result analysis.
% \subsection{Experiments setup}
\subsection{The illustration for data augmentation}\label{app_diag}
\begin{figure}[h]
    \centering
    \vspace{-10pt}
    % \subfigure[The adaptive data augmentation pipeline]{
    %     \includegraphics[width=0.6\columnwidth]{figs/adaug.pdf}
    %     \label{fig:adaug}
    %     % \caption{}
    % }
    \begin{minipage}[c]{0.25\columnwidth}
    \centering
        \includegraphics[width=1\columnwidth]{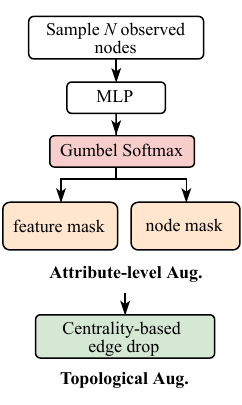}
        \label{fig:adaug}
        
        {(a) The adaptive data augmentation pipeline} 
    \end{minipage}
    % \hfill
    % \hspace{10pt}
    % \subfigure[Illustration of obtaining 3 samples after attribute-level and topological augmentations with $N=3$. In the training stage, $N_o=6$ .]{
    %     \includegraphics[width=\columnwidth]{figs/data_aug_details.pdf}
    %     \label{fig:data_aug}
    %     % \caption{}
    %     }
    \begin{minipage}[c]{0.7\columnwidth}
    \centering
    % \vspace{7pt}
        \includegraphics[width=1\columnwidth]{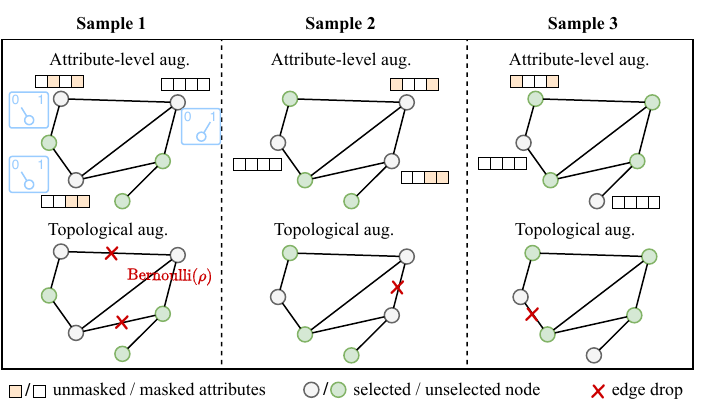}
        \label{fig:data_aug}
        \\
        {(b) Illustration of obtaining 3 samples after attribute-level and topological augmentations with $N=3$. In the training stage, $N_o=6$.}
    \end{minipage}
    \caption{The diagram of adaptive data augmentation module.}
     \label{fig:diag_data_aug}  
     % \vspace{-40pt}
\end{figure}

As shown in Fig. \ref{fig:diag_data_aug}, we further diagram the proposed data augmentation module. Specifically, among the observed nodes, we first randomly select $N$ nodes while keeping the remaining nodes unchanged. Then, attribute-level and
topological augmentations are sequentially applied to each
selected node (shown in Fig. \ref{fig:diag_data_aug}(a)). In Fig.  \ref{fig:diag_data_aug}(b), we showcase the internal modifications on three samples with augmentations.

\subsection{Proof of Theorem 3.1}\label{app_diag}

For the topologically augmented graphon $W'$, there is $W' = (\mathbbm{1} - \Phi) \odot W$. Recall that the definition of homomorphism density $t$ is $t(\mathcal{G}, W)=\int_{[0,1] V(\mathcal{G})} \prod_{i, j \in E(\mathcal{G})} W\left(x_i, x_j\right) \prod_{i \in V(\mathcal{G})} d x_i$. 
And the homomorphism density of $W'$ is
\begin{equation}
\begin{aligned}
t(\mathcal{G}, W') &= \int_{[0,1]^{|V(\mathcal{G})|}} \prod_{i, j \in E(\mathcal{G})} W'\left(x_i, x_j\right) \prod_{i \in V(\mathcal{G})} d x_i\\
&= \int_{[0,1] ^{|V(\mathcal{G})|}} \prod_{i, j \in E(\mathcal{G})} ((\mathbbm{1} - \Phi) \odot W)\left(x_i, x_j\right) \prod_{i \in V(\mathcal{G})} d x_i\\
&= \int_{[0,1]^{|V(\mathcal{G})|}} \prod_{i, j \in E(\mathcal{G})}(1 - \Phi_{i,j}) W\left(x_i, x_j\right) \prod_{i \in V(\mathcal{G})} d x_i\\
&= \prod_{i, j \in E(\mathcal{G})}(1 - \Phi_{i,j}) \int_{[0,1]^{|V(\mathcal{G})|}} \prod_{i, j \in E(\mathcal{G})} W\left(x_i, x_j\right) \prod_{i \in V(\mathcal{G})} d x_i\\
&= \prod_{i, j \in E(\mathcal{G})}(1 - \Phi_{i,j}) t(\mathcal{G}, W)
\end{aligned}
\end{equation}
Since $\Phi_{i,j}$ is determined by sampling under the edge drop probability $\rho$ and irrelevant to $dx$, it can be excluded from the integration. 
Following triangle inequality in G-mixup \cite{han2022g}, 
$\left|t(\mathcal{G}, W') - t(\mathcal{G}, W)\right|\leq e(\mathcal{G})\|W - W'\|_{\square}$. 
Let $\lambda=\prod_{i, j \in E(\mathcal{G})}(1 - \Phi_{i,j})$, and then the upper bound is derived as 
\begin{equation}
    \left|t(\mathcal{G}, W')-t(\mathcal{G}, W)\right| = \left|\lambda t(\mathcal{G}, W) - t(\mathcal{G}, W)\right|\leq e(\mathcal{G})\|(1-\lambda) W\|_{\square} = (1-\lambda) e(\mathcal{G})\| W\|_{\square}
\end{equation}
\subsection{Dataset description}\label{app_exp}
For a broader performance evaluation of the proposed model, we conduct Kriging experiments on three publicly available time series datasets with geographic properties, which are described in Table \ref{tab:data_desc}.
\begin{table}[h]
\centering
\caption{Dataset description}
\label{tab:data_desc}
{
\begin{tabular}{cccc}
\toprule
           & \textbf{Attributes}   & \textbf{Nodes} & \textbf{Resolution} \\\hline
PeMS       & Traffic flow & 325   & 5 min           \\\hline
NREL       & Solar power  & 137   & 5 min           \\\hline
Wind & Wind speed   & 18    & hourly    \\
\bottomrule
\end{tabular}}
\end{table}

1) PeMS: The dataset aggregates a 5-minute traffic flow across 325 traffic stations, collected from the Caltrans Performance Measurement System over a 2-month period (January 1st, 2017 to March 1st, 2017). 
2) NREL: It records 5-minute solar power output from 137 photovoltaic plants in Alabama in 2006, which is extracted from \citep{wu2021inductive}.
3) Wind: 
This dataset records onshore renewable energy generation for Greece, which contains hourly wind speed aggregation on 18 installations over 4 years (2017-2020) \citep{vartholomaios2021short}.

% \renewcommand\arraystretch{1.2}
% \begin{wraptable}{r}{0.5\textwidth}
% \centering
% \caption{Dataset description}
% \label{tab:data_desc}
% \resizebox{0.49\textwidth}{!}{
% \begin{tabular}{cccc}
% \toprule
%            & \textbf{Attributes}   & \textbf{Nodes} & \textbf{Resolution} \\\hline
% PeMS       & Traffic flow & 325   & 5 min           \\\hline
% NREL       & Solar power  & 137   & 5 min           \\\hline
% Greek wind & Wind speed   & 18    & hourly    \\
% \bottomrule
% \end{tabular}}
% \end{wraptable}

Gaussian kernel-based inverse distance is employed to calculate the adjacent matrix $\mathbf{A}$ following \citep{li2018dcrnn_traffic}, where road network distance is for the PeMS dataset and Euclidean distance is for NREL and Wind datasets. For nodes $v_i$ and $v_j$, the related entry $\mathbf{A}_{ij}$ in $\mathbf{A}$ is calculated by
\begin{equation}
\mathbf{A}_{i j}=\exp \left(-\left(\frac{\mathrm{dist}\left(v_i, v_j\right)}{\sigma}\right)^2\right),
\end{equation}
where $\sigma$ is the standard deviation of the distance.
% In each dataset, 80\% randomly selected nodes are set as observations (i.e., $\mathcal{V}$) for training and the remaining are regarded as unobserved nodes for testing. The data are normalized by the min-max scaler. 

\subsection{Baselines}\label{app_bsl} %{This part could be put in the appendix} 
We compare seven advanced Kriging models involving statistical models and deep learning-based methods. They are 
1) KNN-IDW: A KNN model which incorporates inverse distance weighted interpolation for Kriging. The weighted average of the $K$ nearest observed nodes to the unobserved node is taken as the estimation, and the weight is associated with the entry in the adjacent matrix. We set $K=5$ in this paper.
2) XGBoost \citep{chen2016xgboost}: We train an extreme gradient boosting model, in which the geolocations are regarded as the input, and the values of the attributes are the output.
3) OKriging \citep{bostan2017basic}: 
Ordinary Kriging. It estimates the unknown nodes with a known variogram under the Gaussian process, which is a typical spatial interpolation model.
4) GPMF \citep{strahl2020scalable}: It is a graph-based prior probabilistic matrix factorization, in which the graph structure is used as the side information to achieve the matrix factorization.
5) BGRL \citep{thakoor2021large}: It is a self-supervised graph representation learning method that learns by predicting alternative augmentations of the input. It uses only simple augmentations without explicit negative samples. The model is also tailored to the Kriging task under the pretraining and then finetuning SSL paradigm.
6) IGNNK \citep{wu2021inductive}: It constructs dynamic subgraphs by random sampling and uses diffusion graph convolutional network \citep{li2018dcrnn_traffic} for the spatiotemporal Kriging.
7) KCN \citep{appleby2020kriging}: A graph convolutional network for Kriging, which makes direct use of $K$ nearest neighboring observations for graph message passing. We also set $K=5$.

For the neural networks, we split the time series attributes in each node by slide window with size $ 24$ for input.

\subsection{Evaluation metrics}\label{app_metric} To quantitatively characterize the performance of Kriging models, mean absolute error (MAE), root mean square error (RMSE), and mean absolute percentage error (MAPE) are adopted to evaluate the estimations $\hat{\mathbf{Y}}$ and ground truth ${\mathbf{Y}}$ on unobserved nodes. The first reflects the average error between $\hat{\mathbf{Y}}$ and ${\mathbf{Y}}$. The second provides higher weights to larger errors, which is more sensitive to outliers. The last one indicates the average percentage deviation of the $\hat{\mathbf{Y}}$ from ${\mathbf{Y}}$. Their definitions are 

\begin{equation}
\mathrm{MAE}=\frac{1}{N_u \times T} \sum_{j=1}^{N_u}\left(\sum_{t=1}^{T}\left|y^t_j-\hat{y}^t_j\right|\right),
\end{equation}

\begin{equation}
\mathrm{RMSE}=\left[\frac{1}{N_u \times T} \sum_{j=1}^{N_u}\left(\sum_{t=1}^{T}\left(y^t_j-\hat{y}^t_j\right)^2\right)\right]^{1 / 2},
\end{equation}

\begin{equation}
\mathrm{MAPE}=\frac{1}{N_u \times T} \sum_{j=1}^{N_u}\left(\sum_{t=1}^{T}\frac{\left|y^t_j-\hat{y}^t_j\right|}{y^t_j}\right),
\end{equation}
where $N_u$ denotes the number of unobserved nodes. $T$ is the time horizon to perform Kriging. $y^t_j$ and $\hat{y}^t_j$ are scalars at the time step $t$ on node $v_j$, which indicate ground truth and the  estimation, respectively.

\subsection{Parameter sensitivity}
\begin{figure}[h]
\begin{minipage}[c]{0.3\columnwidth}
% \hspace{30pt}
    \centering
    \includegraphics[width=\columnwidth]{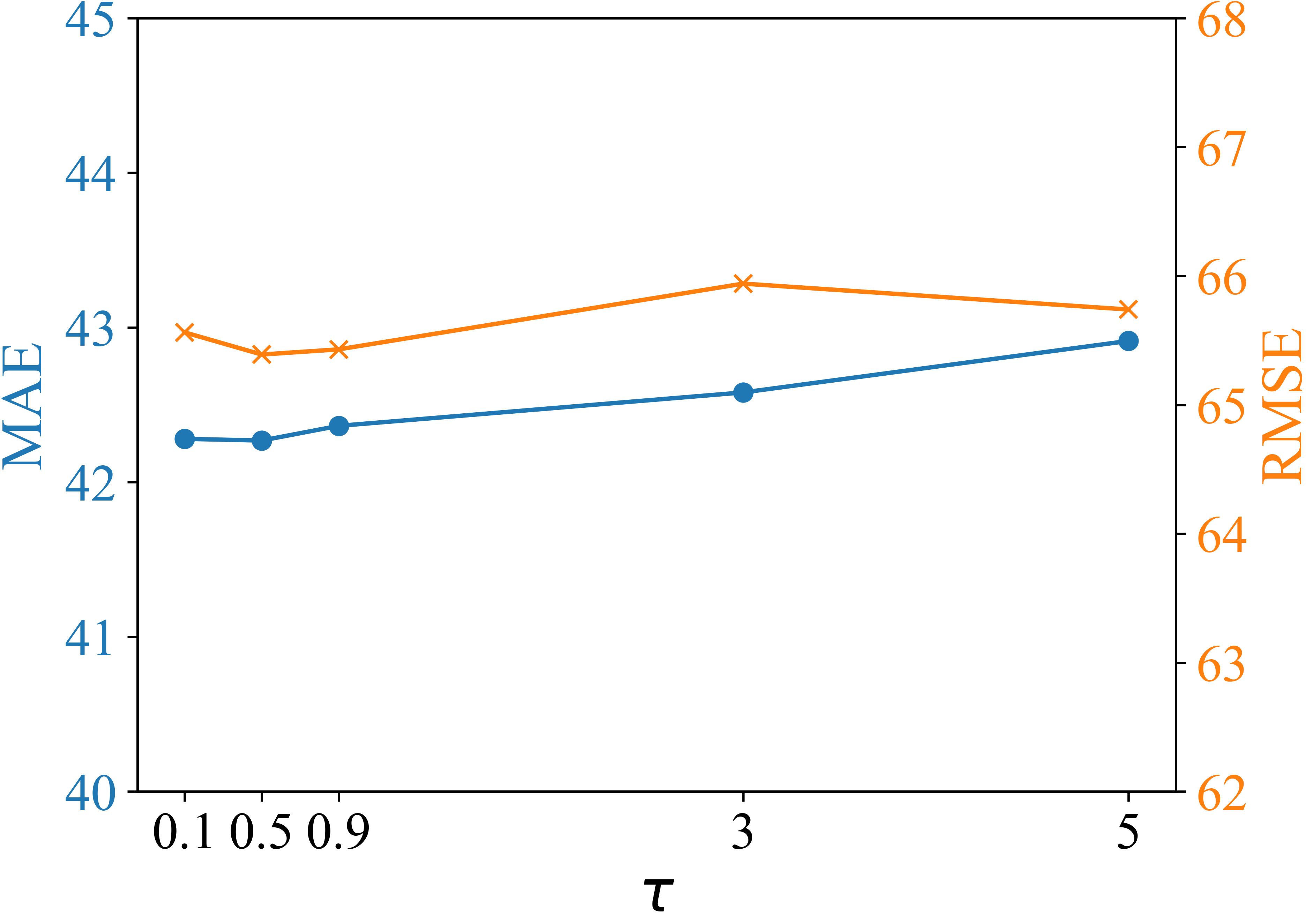}
    \\{(a) The performance under different $\tau$}
    \vspace{10pt}
\end{minipage}
% \hspace{-10pt}
\hfill
\begin{minipage}[c]{0.3\columnwidth}
    \centering
    \includegraphics[width=\columnwidth]{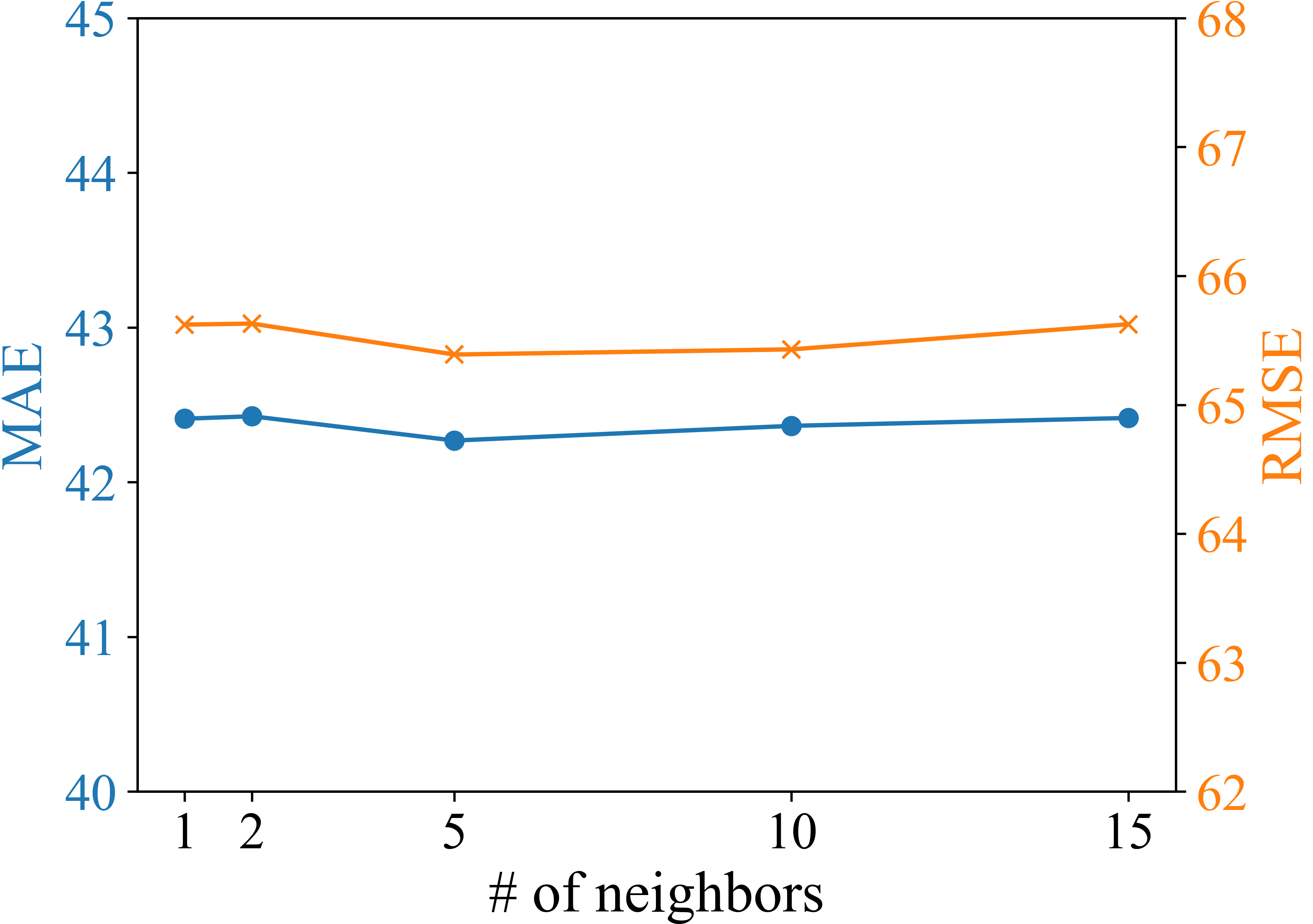}
    % \hspace{30pt}
    \\{(b) The performance under different $k$}
    \vspace{10pt}
\end{minipage}
\hfill
\begin{minipage}[c]{0.3\columnwidth}
% \hspace{30pt}
    \centering
    \includegraphics[width=\columnwidth]{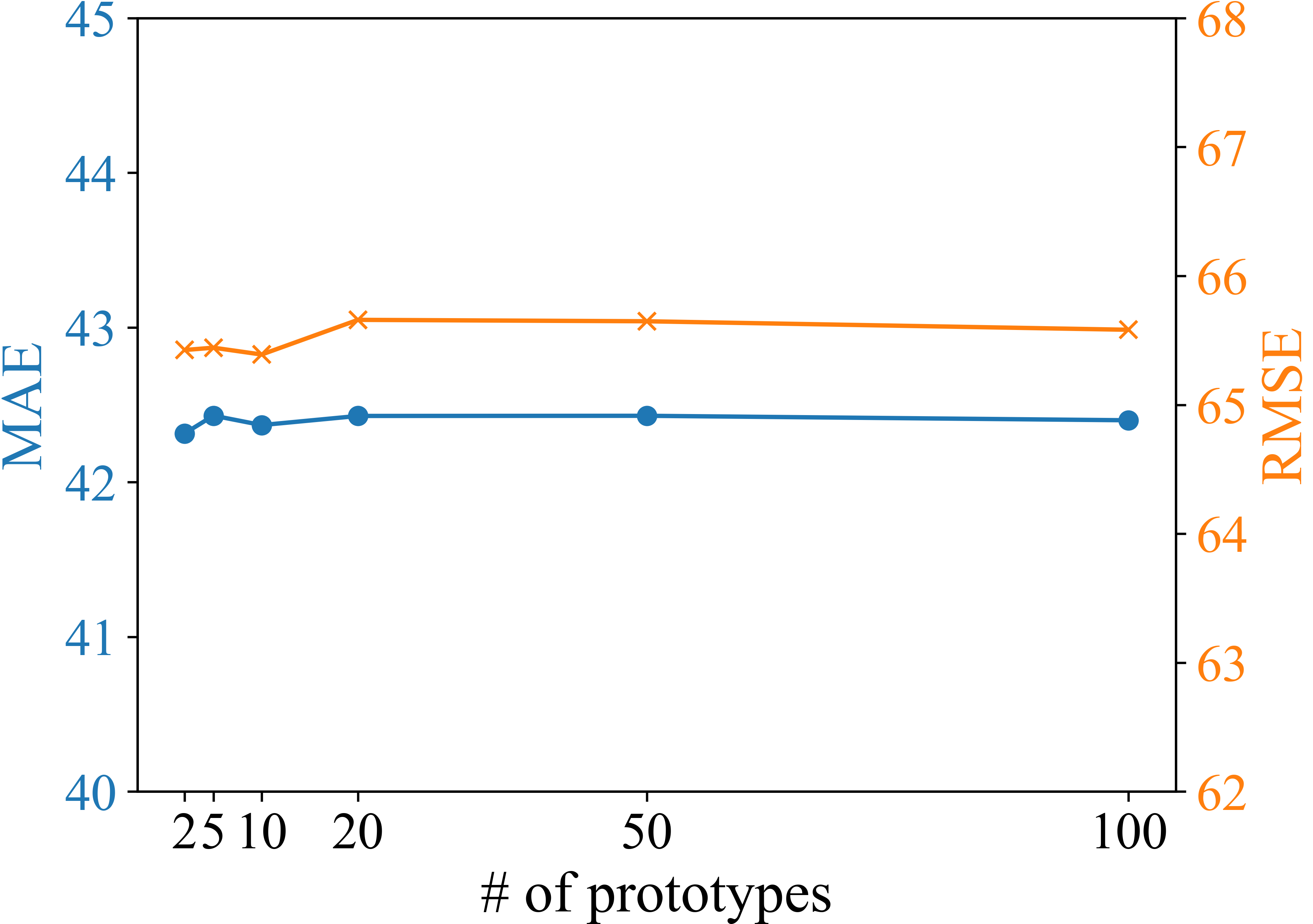}
    \\{(c) The performance under different $H$}
    \vspace{10pt}
\end{minipage}
\caption{Analysis of the hyper-parameters sensitivity in PeMS dataset}
\label{fig:tau}
\end{figure}

To examine whether the proposed method is robust to the hyper-parameters, as shown in Fig. \ref{fig:tau}, we investigate the sensitivity of temperature $\tau$ in eq. (3) in main text, numbers of top-$k$ neighbors, and numbers of prototypes $H$ in the PeMS dataset under $\rm MAE$ and $\rm RMSE$ metrics. 

As can be seen, the best performance is obtained when $\tau = 0.5,k=5,H=10$.

\subsection{Spatial visualization of Kriging results} 
\begin{figure}[h]
    \centering
    \includegraphics[width=0.8\columnwidth]{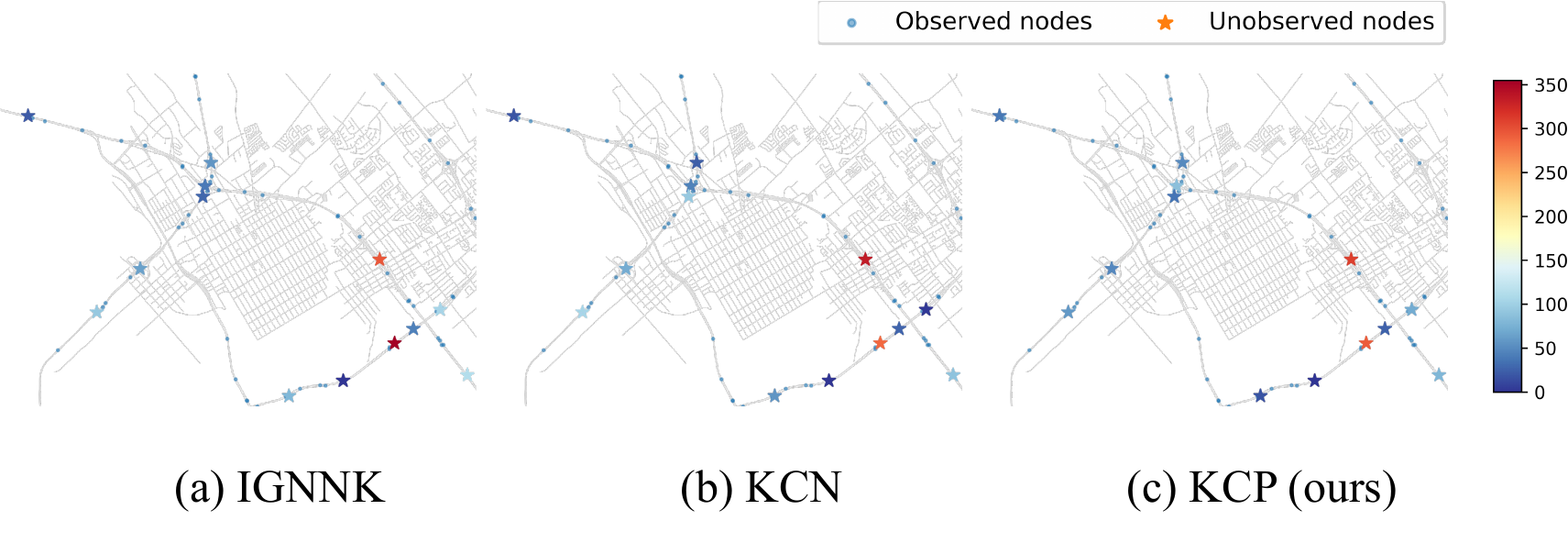}
    \caption{The visualization of Kriging errors at a particular time point on the PeMS. The stars represent the unknown nodes and the dot points mean the observed nodes.}
    \label{fig:spatial_vis}
\end{figure}

Based on PeMS, we further provide intuitive visualization about the absolute Kriging errors across the geographic locations, which are calculated by $|estimation-ground~truth|$ and illustrated in Fig. \ref{fig:spatial_vis}.
For demonstration, we choose the traffic flow of the nodes at a particular time and show the Kriging errors from the graph-based baselines and our method, respectively. It can be seen that the errors in our model are the smallest, which also indicates its superiority.

\end{document}